\documentclass[10pt,twocolumn,letterpaper]{article}

\usepackage[pagenumbers]{cvpr} %

\usepackage{graphicx}
\usepackage{amsmath}
\usepackage{amssymb}
\usepackage{booktabs}
\usepackage[all]{nowidow}
\usepackage{verbatim}
\usepackage[dvipsnames]{xcolor}
\usepackage{paralist}
\newcommand{\link}[1]{{\color{blue}\href{#1}{#1}}}

\DeclareMathOperator*{\argmax}{arg\,max}

\usepackage[pagebackref,breaklinks,colorlinks]{hyperref}

\usepackage[capitalize]{cleveref}
\crefname{section}{Sec.}{Secs.}
\Crefname{section}{Section}{Sections}
\Crefname{table}{Table}{Tables}
\crefname{table}{Tab.}{Tabs.}

\def\cvprPaperID{3954} %
\def\confName{CVPR}
\def\confYear{2023}

\begin{document}

\title{Invertible Neural Skinning}

\author{
  Yash Kant$^{1,2}$, Aliaksandr Siarohin$^{2}$, Riza Alp Guler$^{2}$, Menglei Chai$^{2}$, Jian Ren$^{2}$, \\ Sergey Tulyakov$^{2}$, Igor Gilitschenski$^{1}$ \\
  $^1$University of Toronto, $^2$Snap Research \\
  \link{https://yashkant.github.io/invertible-neural-skinning/}
  }
\maketitle

\newcommand{\dpose}{\boldsymbol{\theta}^t_{d}}
\newcommand{\cpose}{\boldsymbol{\theta}_{c}}
\newcommand{\cdpose}{\boldsymbol{\theta}^t_{c\rightarrow d}}
\newcommand{\dcpose}{\boldsymbol{\theta}^t_{d\rightarrow c}}

\newcommand{\cpoint}{\mathbf{p}_{c}}
\newcommand{\dpoint}{\mathbf{p}^t_{d}}
\newcommand{\qdpoint}{\mathbf{q}^t_{d}}
\newcommand{\qcpoint}{\mathbf{q}_{c}}
\newcommand{\bpose}{\boldsymbol{\theta}^{t}}
\newcommand{\pose}{\boldsymbol{\theta}}
\newcommand{\innd}{\mathbf{H}_{d}}
\newcommand{\innc}{\mathbf{H}_{c}}

\newcommand{\pin}{PIN}
\newcommand{\ins}{INS}
\newcommand{\var}{\texttt}

\newcommand{\pardev}[2]{\frac{\partial {#1}}{\partial {#2}}}

\newcommand{\netparam}{\sigma}

\newcommand{\latent}{\mathbf{z}}
\newcommand{\spoint}{\hat{\mathbf{x}}}
\newcommand{\function}[1]{{#1}}
\newcommand{\mfunction}[1]{\mathbf{#1}}

\newcommand{\raydir}{\mathbf{r}_d}
\newcommand{\rayori}{\mathbf{r}_c}

\newcommand{\pixel}{\mathbf{p}}
\newcommand{\loss}{\mathcal{L}}
\newcommand{\bodypose}{\boldsymbol{\theta}}
\newcommand{\bodyshape}{\boldsymbol{\beta}}

\newcommand{\bone}{\boldsymbol{B}}
\newcommand{\nbone}{n_{\boldsymbol{B}}}

\newcommand{\occ}{o}
\newcommand{\tex}{\mathbf{t}}
\newcommand{\norm}{\mathbf{n}}
\newcommand{\skin}{\mathbf{w}}
\newcommand{\warp}{\mathbf{d}}

\newcommand{\fdef}{\mathcal{D}}
\newcommand{\focc}{\mathcal{O}}
\newcommand{\ftex}{\mathcal{T}}
\newcommand{\fnorm}{\mathcal{N}}
\newcommand{\fskin}{\mathcal{W}}
\newcommand{\fwarp}{\mathcal{M}}

\newcommand{\zocc}{\latent_\text{shape}}
\newcommand{\ztex}{\latent_\tex}
\newcommand{\znorm}{\latent_\text{detail}}
\newcommand{\zskin}{\latent_\text{shape}}
\newcommand{\zwarp}{\latent_\warp}

\newcommand{\feat}{\mathbf{f}}

\newcommand{\R}[1]{\mathbb{R}^{#1}}

\newcommand{\gt}{}
\newcommand{\jac}{\mathbf{J}}

\definecolor{best_color}{HTML}{FCE5CD}
\definecolor{better_color}{HTML}{DEEDF2}

\newcommand{\pt}{\mathbf{x}} 
\newcommand{\cnlpt}{\pt_c}
\newcommand{\obspt}{\pt_o}
\newcommand{\ray}{\mathbf{r}} 
\newcommand{\cam}{\mathbf{e}} 
\newcommand{\nroffset}{{\Delta}x, {\Delta}y, {\Delta}z}
\newcommand{\nroffsetpack}{\Delta \pt}

\newcommand{\cnlvolfunc}{F_c}
\newcommand{\obsvolfunc}{F_o} 
\newcommand{\posecorrectfunc}{P_{\rm pose}}

\newcommand{\motionfield}{T}
\newcommand{\skelmotionfield}{T_{\rm skel}}
\newcommand{\nrmotionfield}{T_{\rm NR}}

\newcommand{\rotbasis}{R}
\newcommand{\transbasis}{\mathbf{t}}

\newcommand{\weightcnl}{w_c}
\newcommand{\weightvolcnl}{W_c}
\newcommand{\weightobs}{w_o}
\newcommand{\weightvolobs}{W_o}
\newcommand{\wightvoldelta}{\Delta \weightvolcnl}
\newcommand{\weightvolgaussian}{W_g}

\newcommand{\weightcnn}{\rm CNN} 
\newcommand{\weightcnnlatent}{\textbf{z}}

\newcommand{\mlp}{\rm MLP} 
\newcommand{\posencode}{\gamma} 
\newcommand{\mlpcolor}{\mathbf{c}}
\newcommand{\mlpdensity}{\mathbf{\sigma}}
\newcommand{\mlpalpha}{\mathbf{\alpha}}

\newcommand{\joints}{J}
\newcommand{\joint}{j}
\newcommand{\jangle}{\boldsymbol{\omega}}
\newcommand{\jangles}{\Omega}
\newcommand{\spacetfm}{M}

\newcommand{\volumerender}{\Gamma} 
\newcommand{\fglikelihood}{f}

\newcommand{\skelparam}{\theta_{\text{skel}}}
\newcommand{\nrparam}{\theta_{\rm NR}} 
\newcommand{\appearanceparam}{\theta_c} 
\newcommand{\posecorrectparam}{\theta_{\rm pose}}

\newcommand{\allparam}{\Theta}
\newcommand{\allparamdetail}{\appearanceparam, \skelparam, \nrparam, \posecorrectparam}
\newcommand{\ba}{\mathbf{a}}\newcommand{\bA}{\mathbf{A}}
\newcommand{\bb}{\mathbf{b}}\newcommand{\bB}{\mathbf{B}}
\newcommand{\bc}{\mathbf{c}}\newcommand{\bC}{\mathbf{C}}
\newcommand{\bd}{\mathbf{d}}\newcommand{\bD}{\mathbf{D}}
\newcommand{\be}{\mathbf{e}}\newcommand{\bE}{\mathbf{E}}
\newcommand{\bff}{\mathbf{f}}\newcommand{\bF}{\mathbf{F}} %
\newcommand{\bg}{\mathbf{g}}\newcommand{\bG}{\mathbf{G}}
\newcommand{\bh}{\mathbf{h}}\newcommand{\bH}{\mathbf{H}}
\newcommand{\bi}{\mathbf{i}}\newcommand{\bI}{\mathbf{I}}
\newcommand{\bj}{\mathbf{j}}\newcommand{\bJ}{\mathbf{J}}
\newcommand{\bk}{\mathbf{k}}\newcommand{\bK}{\mathbf{K}}
\newcommand{\bl}{\mathbf{l}}\newcommand{\bL}{\mathbf{L}}
\newcommand{\bm}{\mathbf{m}}\newcommand{\bM}{\mathbf{M}}
\newcommand{\bn}{\mathbf{n}}\newcommand{\bN}{\mathbf{N}}
\newcommand{\bo}{\mathbf{o}}\newcommand{\bO}{\mathbf{O}}
\newcommand{\bp}{\mathbf{p}}\newcommand{\bP}{\mathbf{P}}
\newcommand{\bq}{\mathbf{q}}\newcommand{\bQ}{\mathbf{Q}}
\newcommand{\br}{\mathbf{r}}\newcommand{\bR}{\mathbf{R}}
\newcommand{\bs}{\mathbf{s}}\newcommand{\bS}{\mathbf{S}}
\newcommand{\bt}{\mathbf{t}}\newcommand{\bT}{\mathbf{T}}
\newcommand{\bu}{\mathbf{u}}\newcommand{\bU}{\mathbf{U}}
\newcommand{\bv}{\mathbf{v}}\newcommand{\bV}{\mathbf{V}}
\newcommand{\bw}{\mathbf{w}}\newcommand{\bW}{\mathbf{W}}
\newcommand{\bx}{\mathbf{x}}\newcommand{\bX}{\mathbf{X}}
\newcommand{\by}{\mathbf{y}}\newcommand{\bY}{\mathbf{Y}}
\newcommand{\bz}{\mathbf{z}}\newcommand{\bZ}{\mathbf{Z}}

\newcommand{\balpha}{\boldsymbol{\alpha}}\newcommand{\bAlpha}{\boldsymbol{\Alpha}}
\newcommand{\bbeta}{\boldsymbol{\beta}}\newcommand{\bBeta}{\boldsymbol{\Beta}}
\newcommand{\bgamma}{\boldsymbol{\gamma}}\newcommand{\bGamma}{\boldsymbol{\Gamma}}
\newcommand{\bdelta}{\boldsymbol{\delta}}\newcommand{\bDelta}{\boldsymbol{\Delta}}
\newcommand{\bepsilon}{\boldsymbol{\epsilon}}\newcommand{\bEpsilon}{\boldsymbol{\Epsilon}}
\newcommand{\bzeta}{\boldsymbol{\zeta}}\newcommand{\bZeta}{\boldsymbol{\Zeta}}
\newcommand{\beeta}{\boldsymbol{\eta}}\newcommand{\bEta}{\boldsymbol{\Eta}} %
\newcommand{\btheta}{\boldsymbol{\theta}}\newcommand{\bTheta}{\boldsymbol{\Theta}}
\newcommand{\biota}{\boldsymbol{\iota}}\newcommand{\bIota}{\boldsymbol{\Iota}}
\newcommand{\bkappa}{\boldsymbol{\kappa}}\newcommand{\bKappa}{\boldsymbol{\Kappa}}
\newcommand{\blambda}{\boldsymbol{\lambda}}\newcommand{\bLambda}{\boldsymbol{\Lambda}}
\newcommand{\bmu}{\boldsymbol{\mu}}\newcommand{\bMu}{\boldsymbol{\Mu}}
\newcommand{\bnu}{\boldsymbol{\nu}}\newcommand{\bNu}{\boldsymbol{\Nu}}
\newcommand{\bxi}{\boldsymbol{\xi}}\newcommand{\bXi}{\boldsymbol{\Xi}}
\newcommand{\bomikron}{\boldsymbol{\omikron}}\newcommand{\bOmikron}{\boldsymbol{\Omikron}}
\newcommand{\bpi}{\boldsymbol{\pi}}\newcommand{\bPi}{\boldsymbol{\Pi}}
\newcommand{\brho}{\boldsymbol{\rho}}\newcommand{\bRho}{\boldsymbol{\Rho}}
\newcommand{\bsigma}{\boldsymbol{\sigma}}\newcommand{\bSigma}{\boldsymbol{\Sigma}}
\newcommand{\btau}{\boldsymbol{\tau}}\newcommand{\bTau}{\boldsymbol{\Tau}}
\newcommand{\bypsilon}{\boldsymbol{\ypsilon}}\newcommand{\bYpsilon}{\boldsymbol{\Ypsilon}}
\newcommand{\bphi}{\boldsymbol{\phi}}\newcommand{\bPhi}{\boldsymbol{\Phi}}
\newcommand{\bchi}{\boldsymbol{\chi}}\newcommand{\bChi}{\boldsymbol{\Chi}}
\newcommand{\bpsi}{\boldsymbol{\psi}}\newcommand{\bPsi}{\boldsymbol{\Psi}}
\newcommand{\bomega}{\boldsymbol{\omega}}\newcommand{\bOmega}{\boldsymbol{\Omega}}

\newcommand{\nA}{\mathbb{A}}
\newcommand{\nB}{\mathbb{B}}
\newcommand{\nC}{\mathbb{C}}
\newcommand{\nD}{\mathbb{D}}
\newcommand{\nE}{\mathbb{E}}
\newcommand{\nF}{\mathbb{F}}
\newcommand{\nG}{\mathbb{G}}
\newcommand{\nH}{\mathbb{H}}
\newcommand{\nI}{\mathbb{I}}
\newcommand{\nJ}{\mathbb{J}}
\newcommand{\nK}{\mathbb{K}}
\newcommand{\nL}{\mathbb{L}}
\newcommand{\nM}{\mathbb{M}}
\newcommand{\nN}{\mathbb{N}}
\newcommand{\nO}{\mathbb{O}}
\newcommand{\nP}{\mathbb{P}}
\newcommand{\nQ}{\mathbb{Q}}
\newcommand{\nR}{\mathbb{R}}
\newcommand{\nS}{\mathbb{S}}
\newcommand{\nT}{\mathbb{T}}
\newcommand{\nU}{\mathbb{U}}
\newcommand{\nV}{\mathbb{V}}
\newcommand{\nW}{\mathbb{W}}
\newcommand{\nX}{\mathbb{X}}
\newcommand{\nY}{\mathbb{Y}}
\newcommand{\nZ}{\mathbb{Z}}

\newcommand{\cA}{\mathcal{A}}
\newcommand{\cB}{\mathcal{B}}
\newcommand{\cC}{\mathcal{C}}
\newcommand{\cD}{\mathcal{D}}
\newcommand{\cE}{\mathcal{E}}
\newcommand{\cF}{\mathcal{F}}
\newcommand{\cG}{\mathcal{G}}
\newcommand{\cH}{\mathcal{H}}
\newcommand{\cI}{\mathcal{I}}
\newcommand{\cJ}{\mathcal{J}}
\newcommand{\cK}{\mathcal{K}}
\newcommand{\cL}{\mathcal{L}}
\newcommand{\cM}{\mathcal{M}}
\newcommand{\cN}{\mathcal{N}}
\newcommand{\cO}{\mathcal{O}}
\newcommand{\cP}{\mathcal{P}}
\newcommand{\cQ}{\mathcal{Q}}
\newcommand{\cR}{\mathcal{R}}
\newcommand{\cS}{\mathcal{S}}
\newcommand{\cT}{\mathcal{T}}
\newcommand{\cU}{\mathcal{U}}
\newcommand{\cV}{\mathcal{V}}
\newcommand{\cW}{\mathcal{W}}
\newcommand{\cX}{\mathcal{X}}
\newcommand{\cY}{\mathcal{Y}}
\newcommand{\cZ}{\mathcal{Z}}

\newcommand{\figref}[1]{Fig.~\ref{#1}}
\newcommand{\secref}[1]{Section~\ref{#1}}
\newcommand{\algref}[1]{Algorithm~\ref{#1}}
\newcommand{\eqnref}[1]{Eq.~\eqref{#1}}
\newcommand{\tabref}[1]{Tab.~\ref{#1}}

\newcommand{\snarf}{SNARF}
\newcommand{\cadex}{CaDeX}

\def\mc{\mathcal}
\def\mb{\mathbf}

\newcommand{\T}{^{\raisemath{-1pt}{\mathsf{T}}}}

\newcommand{\Perp}{\perp\!\!\! \perp}

\makeatletter
\DeclareRobustCommand\onedot{\futurelet\@let@token\@onedot}
\def\@onedot{\ifx\@let@token.\else.\null\fi\xspace}
\def\eg{e.g\onedot} \def\Eg{E.g\onedot}
\def\ie{i.e\onedot} \def\Ie{I.e\onedot}
\def\cf{cf\onedot} \def\Cf{Cf\onedot}
\def\etc{etc\onedot}
\def\vs{vs\onedot}
\def\wrt{wrt\onedot}
\def\dof{d.o.f\onedot}
\def\etal{et~al\onedot}
\def\iid{i.i.d\onedot}
\makeatother

\renewcommand\UrlFont{\color{blue}\rmfamily}

\newcommand*\rot{\rotatebox{90}}

\newcommand{\boldparagraph}[1]{\vspace{0.2cm}\noindent{\bf #1:}}

\definecolor{turquoise}{cmyk}{0.65,0,0.1,0.3}
\definecolor{purple}{rgb}{0.65,0,0.65}
\definecolor{dark_green}{rgb}{0, 0.5, 0}
\definecolor{orange}{rgb}{0.8, 0.6, 0.2}
\definecolor{red}{rgb}{0.8, 0.2, 0.2}
\definecolor{darkred}{rgb}{0.6, 0.1, 0.05}
\definecolor{blueish}{rgb}{0.0, 0.3, .6}
\definecolor{light_gray}{rgb}{0.7, 0.7, .7}
\definecolor{pink}{rgb}{1, 0, 1}
\definecolor{greyblue}{rgb}{0.25, 0.25, 1}

\newif\ifshowcomments
\showcommentstrue %

\newcommand{\todo}[1]{\noindent{\color{red}{\bf TODO:} {#1}}}

\ifshowcomments

    \newcommand{\oh}[1]{ \noindent {\color{red} {\bf OH:} {#1}} }
    \newcommand{\ag}[1]{ \noindent {\color{red} {\bf AG:} {#1}} }
    \newcommand{\mjb}[1]{ \noindent {\color{red} {\bf MJB:} {#1}} }
    \newcommand{\jy}[1]{ \noindent {\color{red} {\bf JY:} {#1}} }
    \newcommand{\xc}[1]{ \noindent {\color{red} {\bf XC:} {#1}} }
 \else
    \newcommand{\oh}[1]{\unskip}
    \newcommand{\ag}[1]{\unskip}
    \newcommand{\mjb}[1]{\unskip}
    \newcommand{\jy}[1]{\unskip}
    \newcommand{\xc}[1]{\unskip}
\fi   

\newcommand{\update}[1]{{\color{black}{#1}}}

\newcommand{\methodname}{gDNA\xspace}
\newcommand{\suppmat}{Sup.~Mat.\xspace}

\newcommand{\rulesep}{\unskip\ \vrule\ }

\begin{abstract}

Building animatable and editable models of clothed humans from raw 3D scans and poses is a challenging problem.
Existing reposing methods suffer from the limited expressiveness of Linear Blend Skinning (LBS), require costly mesh extraction to generate each new pose, and typically do not preserve surface correspondences across different poses. 
In this work, we introduce Invertible Neural Skinning (INS) to address these shortcomings. 
To maintain correspondences, we propose a Pose-conditioned Invertible Network (PIN) architecture, which extends the LBS process by learning additional pose-varying deformations. 
Next, we combine PIN with a differentiable LBS module to build an expressive and end-to-end Invertible Neural Skinning (INS) pipeline. 
We demonstrate the strong performance of our method by outperforming the state-of-the-art reposing techniques on clothed humans and preserving surface correspondences, while being an order of magnitude faster. 
We also perform an ablation study, which shows the usefulness of our pose-conditioning formulation, and our qualitative results display that INS can rectify artefacts introduced by LBS well.

\end{abstract}

\section{Introduction}
    
    Being able to create animatable representations of clothed humans beyond skinned meshes is essential for building realistic augmented or virtual reality experiences and improving simulators. Towards this goal, we consider the task of building animatable human representations from raw 3D scans and corresponding poses.  
    Prior work in this area has seen a shift from building parametric models of humans \cite{SMPL:2015, Bogo:ECCV:2016, Guler_2019_CVPR}, to more recent works learning implicit 3D neural representations \cite{deng2019neural, saito2019pifu, saito2020pifuhd, Su2022DANBODA, SMPLicit:2021, metaavatar, alldieck2021imghum} from data in canonical space. These canonical representations are animated to a new pose by a learning skinning weight field around them~\cite{snarf, nasa, scanimate, ARAH:ECCV:2022, LEAP:CVPR:21} and applying Linear Blend Skinning (LBS) to warp the surface, where the pose is defined by a bone skeleton underlying the 3D surface. 
    
    \begin{figure}
    \centering
    \includegraphics[width=\columnwidth]{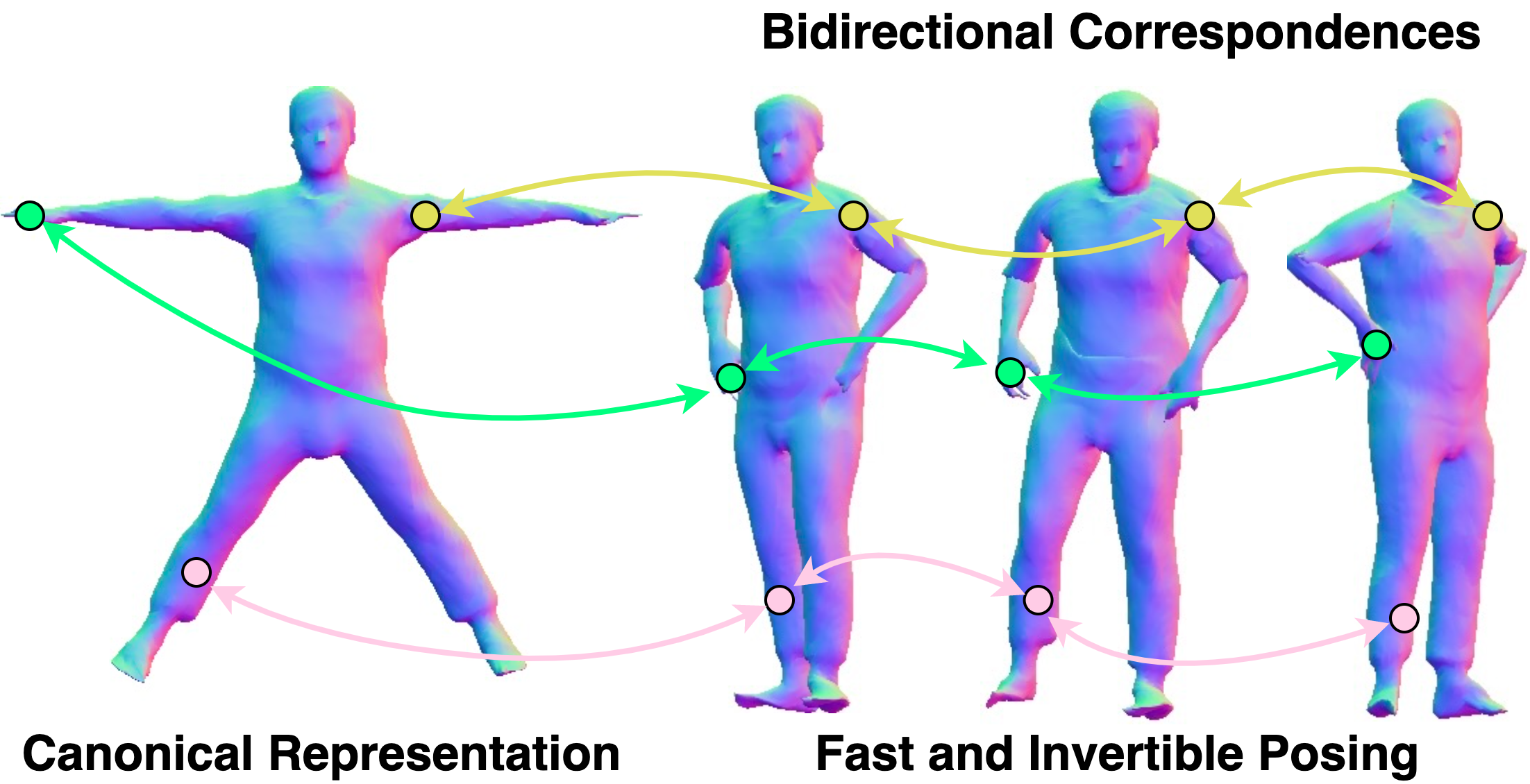}
    \captionof{figure}{\textbf{Fast and Invertible Posing.} We propose an end-to-end learnable reposing pipeline that allows animating implicit surfaces with intricate pose-varying effects, without requiring mesh extraction~\cite{mescheder2019occupancy} for each pose, while also maintaining correspondences across poses.}
    \label{fig:teaser}
\end{figure}

    These prior works generally suffer from the limited expressivity of LBS when handling complex pose-varying deformations, such as those of loose clothes and body tissue (\ie muscle bulges, skin wrinkles). In parametric models like SMPL~\cite{SMPL:2015}, such deformations are represented by adding simple linear pose correctives (aka blend shapes), but these are restrictive and only work for unclothed humans. Implicit methods, to relieve this issue, learn their canonical representations conditioned on the deformed pose~\cite{snarf,nasa}. However, this conditioning comes with two major drawbacks during reposing. Given the sequence of poses, a new mesh has to be extracted from scratch for each pose, which becomes a bottleneck when animating subjects at a high frame-rate or resolution. Also, as a consequence of this step, correspondences (topology preservation) between the surfaces of the same subject across different poses are lost.  

     Invertible Neural Networks (INN)~\cite{realnvp,kingma2018glow,nice} are bijective functions that can preserve exact correspondences between their input and output spaces, while learning complex non-linear transforms between them. This ability of INNs makes them a suitable candidate for reposing, and in this work, we leverage INNs to build an Invertible Neural Skinning (INS) pipeline. For this, we first build a Pose-conditioned Invertible Network, abbreviated as PIN\footnote{We avoid the abbreviation PINN to avoid confusion with Physics-inspired Neural Networks~\cite{pinn}}, to learn pose-conditioned deformations. Next, to create an end-to-end Invertible Neural Skinning (INS) pipeline, we place two PINs around a differentiable LBS module, and use a pose-free canonical representation. These PINs help capture the non-linear surface deformations of clothes across poses and alleviate the volume loss suffered from the LBS operation. Since our canonical representation remains pose-free, we perform the expensive mesh extraction exactly once, and repose the mesh by simply warping it with the learned LBS and an inverse pass through PINs.      
     
    We demonstrate the strong performance of INS by outperforming the previous state-of-the-art reposing method SNARF~\cite{snarf}. On clothed humans data, we find INS provides an absolute gain of roughly 1\% when compared to SNARF with pose-conditioning, and roughly 6\% compared to SNARF without pose-conditioning. We conduct experiments on much simpler minimally clothed human data and obtain competitive results. We also find INS to be an order of magnitude faster at reposing long sequences. We ablate our INS and demonstrate the effectiveness of our pose-conditioning formulation. Our results clearly show that the proposed INS can correct the LBS artefacts well.

\section{Related Work}

\textbf{Representing Articulate Characters in 3D.} Over the years, a significant amount of prior work for building parametric representations of the human body~\cite{SMPL:2015, Guler_2019_CVPR, SMPL-X:2019, osman2020star, FLAME:SiggraphAsia2017, xu2020ghum, anguelov2005scape} or for specific parts such as hands and faces~\cite{moon2020deephandmesh,romero2017embodied} was developed. Beyond humans, recent work developed parametric animal models~\cite{biggs2018creatures, Zuffi_2019_ICCV, dogsout}. Encouraged by the rapid progress in implicit neural 3D representations~\cite{mildenhall2020nerf,mescheder2019occupancy}, a number of works explored building implicit human representations with and without clothing~\cite{human-nerf,huang2020arch, hnerf,zhao2022avatar, xiu2022icon, shao2022doublefield, ARAH:ECCV:2022,SkiRT:3DV:2022,slrf, zheng2022IMavatar}. Representing characters as implicit functions comes with a cost of time-consuming mesh extraction via Marching Cubes~\cite{lorensen1987marching}.

\textbf{Animating 3D Representations with Poses.}
Parametric models usually define the correspondences between poses, represented as a set of bones, and mesh vertices through Linear Blend Skinning (LBS) weights. These weights provide a soft assignment of vertices to human bones. Thus for animation, these models simply transform the vertices using a linear combination of bone transformations. When the parametric model is not available, these weights need to be discovered. To this end, recent works adopt learning-based solutions for discovering LBS weights~\cite{snarf, gDNA,neuralbody,peng2021animatable,tava,LEAP:CVPR:21, nasa,scanimate}. They usually assume a shared canonical space and learn a canonical LBS weight field, which is used for deforming the body in the novel pose during inference. However, at training time, the character needs to be warped backward from deformed to canonical space, i.e. given deformed points, we need to obtain corresponding canonical points. Thus some works~\cite{scanimate, Burov_2021_ICCV, yang2022banmo, Drobyshev22MP, metaavatar} learn LBS weights separately in deformed and canonical spaces, which could be used for establishing correspondences. These generally require cycle-consistency losses for regularization. Recently, SNARF~\cite{snarf} 
proposed to compute these correspondences by finding the solutions of the LBS equation using an iterative solver. We adopt a similar formulation, to discover the correspondences as well.

However, LBS is often insufficient to capture non-linear deformations of flowy clothes and body tissue (\ie muscle bulges). To mitigate this problem, prior works~\cite{snarf, nasa, human-nerf} condition their canonical representations on the deformed pose. Such conditioning helps to alleviate the shortcomings of pure-LBS deformations, but this comes at a cost following two major limitations: 
\begin{itemize}
    \item  \textbf{Slow Reposing.} To generate a new animation given a sequence of poses across time, these methods extract a separate mesh from scratch for each pose. This becomes a bottleneck if we want to pose the character at a high frame-rate or resolution.   
    \item  \textbf{No Correspondences.} As a consequence of the above step, two completely separate meshes get extracted at each pose with no correspondence between them. 
\end{itemize}

In our work, we address both these limitations by extending pure-LBS formulation with additional Pose-conditioned Invertible Networks (PIN), while using a pose-free canonical representation.
\vspace{10pt}

\textbf{Invertible Neural Networks for 3D Vision}. INNs~\cite{nice, realnvp, kingma2018glow,germain2015made,papamakarios2017masked,behrmann2019invertible, irevnet} were initially designed for tractable density estimation of high-dimensional and generative modeling, a.k.a. Normalizing Flows~\cite{Kobyzev_2021, analyze_inv}. Usually, INNs are built by chaining together multiple conditional \textit{Coupling Layers}~\cite{nice,realnvp}, where a single coupling layer defines an invertible transformation between its input and output. The main idea behind \textit{Coupling Layers} is that if we split the input into two parts and only modify the first part while conditioning this modification on the second, this should be trivially invertible. Another popular type of invertible transformations are \textit{Invertible Residual layers}~\cite{irevnet} with small conditioning numbers. They utilize fixed point iterations for finding an inverse. In our work, we mostly rely on \textit{Coupling Layers} since they are faster, and we did not see any additional benefits from \textit{Residual layers}. In the context of 3D vision, INNs were explored for learning primitives of 3D representations~\cite{neuralparts}, doing 3D shape-completion tasks~\cite{Lei2022CaDeX}, and reconstructing dynamic scenes~\cite{Cai2022NDR}. However, to the best of our knowledge, their usage has not been explored for animating 3D characters.

\section{Method}

\textbf{Task Setup.} The goal of our work is to learn a human 3D representation that allows the generation of novel poses beyond original training data (a.k.a. reposing). For each subject, we assume the availability of $N$ pairs consisting of bone poses and 3d meshes denoted as $(\btheta^t, \bM^t)_{t=1}^N$. Such data can be obtained from human scans, and the poses can be estimated by fitting a parametric SMPL-like body model to these scans. \textit{Given this data, we wish to learn a subject-specific implicit neural representation in a canonical space and a method to animate this representation}.   

\textbf{Deformed and Canonical Spaces.} We denote an input point in deformed space as $\dpoint \in \mathbb R^3$  and a point in the canonical space as $\cpoint \in \mathbb R^3$.
Since our input consists of a sequence of deformed (posed) meshes, we use the superscript $t$ to indicate the time-step of capture. As our canonical space is independent of the pose, it is shared across all the time steps; hence, $\cpoint$ is not time-indexed. 

\textbf{Deformed and Canonical Poses.} 
We follow the SMPL model~\cite{SMPL:2015}, which represents body pose as a set of bones in a kinematic tree. While reposing, as we only require the relative pose between canonical and deformed space at any given time $t$, we represent this by $\bodypose^{t} = [ \bone_1,\dots,\bone_{n_b} ]$, where $ \bone_i = [\mathbf{R}_{i}|\mathbf{t}_{i}]$ represents a transformation of the $i^{th}$ bone in 3D space, i.e. $\bB_i\in\mathrm{SE}(3)$ with corresponding rotation $\bR_i\in\nR^{3\times3}$ and translation $\bt_i\in\nR^2$. We denote the total number of bones by $n_b$.

\begin{figure}
    \centering
    \includegraphics[width=0.5\textwidth]{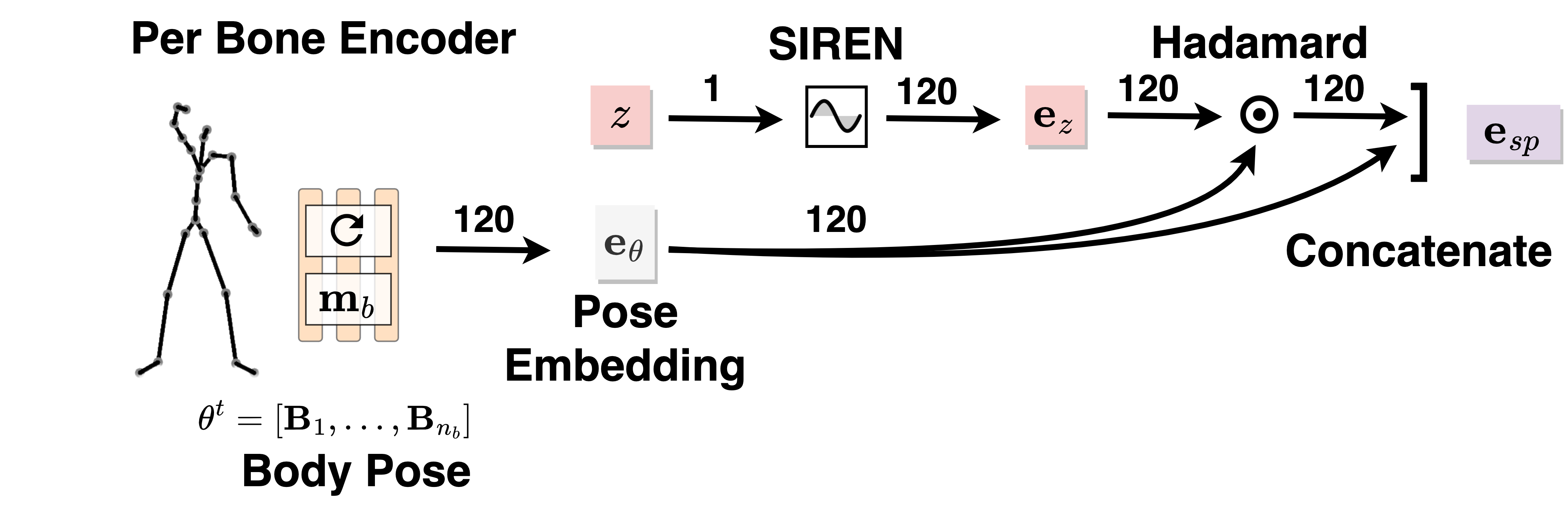}
    \captionof{figure}{\textbf{Space and Pose Aware Conditioning}. We encode the body pose using a per-bone MLP network operating on individual bone transforms. The pose embedding is then fused with space embedding to generate conditioning for PIN.}
    \label{fig:sp_cond}
\end{figure}

\textbf{Pose-free Canonical Occupancy.} To represent a specific subject, we use an Occupancy Network $\mathbf{O}$~\cite{mescheder2019occupancy} conditioned solely on the input point $\cpoint$. The canonical surface $\mathcal{S}_c$ is then represented implicitly as a level-set ($\sigma=0.5$) of this occupancy network. 
\begin{align}
\label{eq:canonical_shape}
\mathcal{S}_c = \{ \cpoint \mid \mathbf{O}(\cpoint) = \sigma\}~~\text{and}~~\mathbf{O}: \mathbb{R}^3\rightarrow [0,1].
\end{align}

To extract this canonical iso-surface as a mesh, we use the MISE~\cite{mescheder2019occupancy} algorithm. This is different from previous works~\cite{snarf, nasa} that use additional pose-conditioning in the canonical occupancy network.

\textbf{Sampling Points.} For both training and evaluation of INS, we sample 3D points in deformed space and get their ground-truth occupancy values of zero or one based on whether they lie outside or the mesh (scan). We put exact details on this sampling in Appendix~\ref*{a:sampling}.

\subsection{Differentiable Forward Blend Skinning}
\label{ssec:difflbs}
To animate our subject from their canonical to deformed pose we use Linear Blend Skinning (LBS), which involves deforming the canonical surface according to a convex combination of rigid bone transforms. Specifically, we use the differentiable LBS formulation from SNARF~\cite{snarf} and summarize it below.

\textbf{Canonical Weight Field.} 
We define a learnable weight field in canonical space parameterized by a neural network, $\mathbf{w}_{lbs}: \mathbb{R}^3 \rightarrow \mathbb{R}^{n_b}$. For a given point in canonical space, this weight field predicts the blend weights corresponding to each bone:
\begin{align}
    \mathbf{w}_{lbs}(\qcpoint) = [w_1, ... , w_{n_b}]~~\text{and}~~w_i \in \mathbb{R}.
\end{align}
To make weights ($w_i$) convex for LBS, they are constrained to be always non-negative and sum to 1 using softmax. 

\textbf{LBS.} Given the above weight field and the relative body pose as bone transforms $\boldsymbol{\theta}^{t}=[\bone_1,\dots,\bone_{n_b}]$, we can forward warp any point $\qcpoint$ of our canonical space to deformed space using Linear Blend Skinning as follows: 
\begin{align}
\qdpoint = \mathbf{lbs}(\mathbf{w}_{lbs}, \qcpoint, \btheta^t) = \left[ \sum_{i=1}^{n_b} \mathbf{w}_{lbs,i}(\qcpoint) \cdot \bone_i \right] \cdot \qcpoint
\label{equ:lbs}
\end{align}
where $\qdpoint$ represents the corresponding point in deformed space where $\qcpoint$ lands after LBS.
 
\begin{figure}
    \centering
    \includegraphics[width=0.45\textwidth]{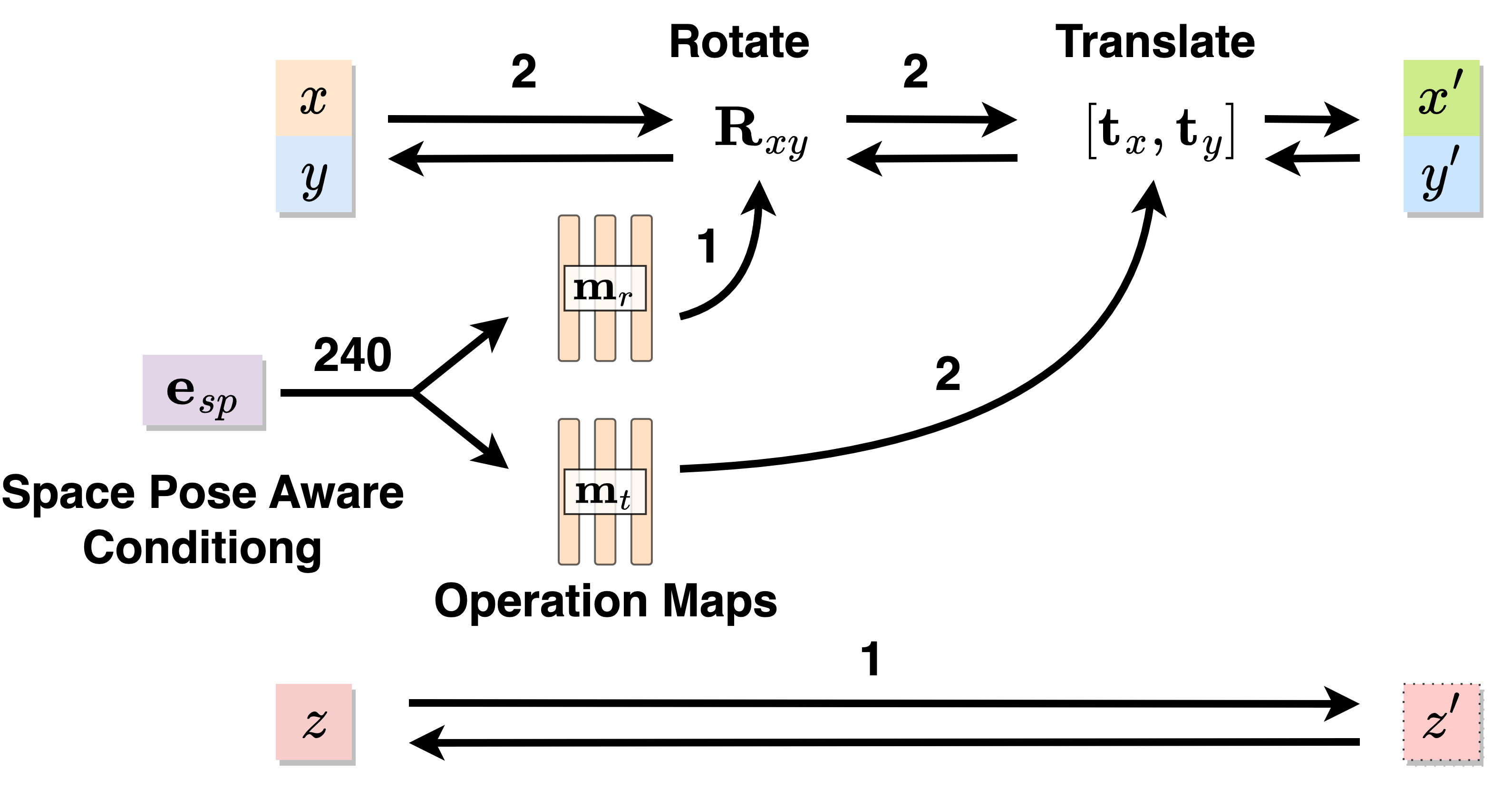}
    \captionof{figure}{\textbf{Pose-conditioned 2D Coupling Layer.} We use the space-pose conditioning to predict the operation parameters using two operation maps (MLPs), and use them to rotate and translate the input split $[x,y]$. In this case, $[z]$ remains unchanged.}
    \label{fig:coupling}
\end{figure}

\begin{figure*}[t!]
    \centering    
    \includegraphics[width=\linewidth]{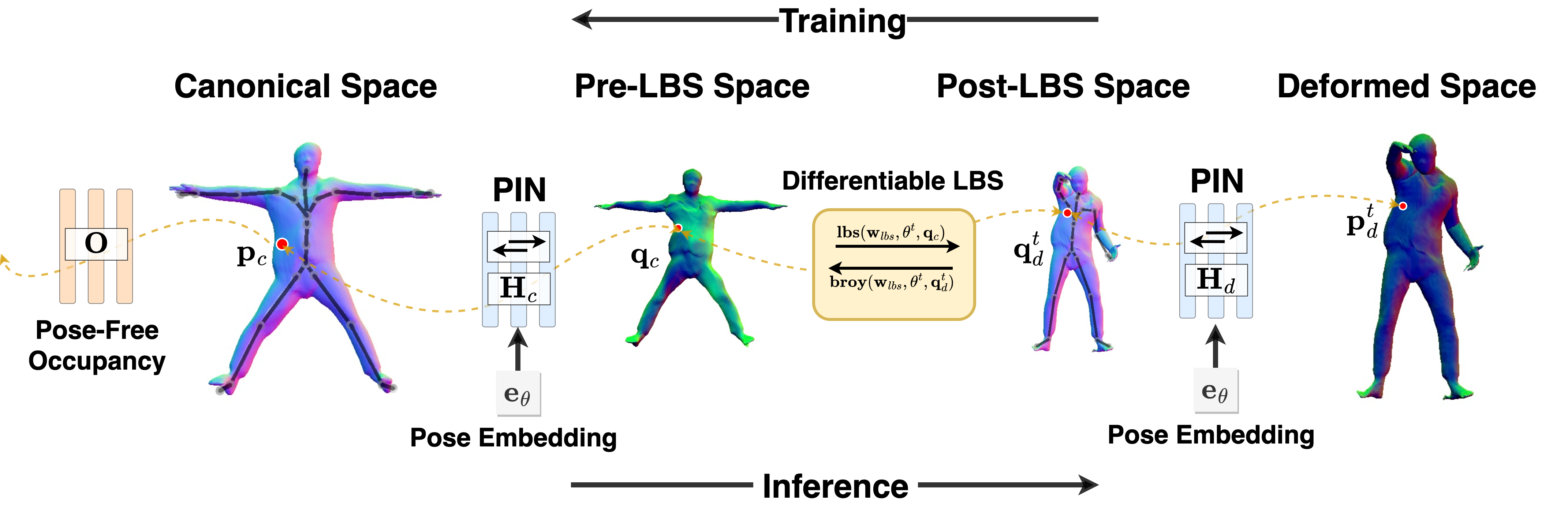}
    \captionof{figure}{\textbf{Invertible Neural Skinning}. Our end-to-end differentiable reposing pipeline consists of two Pose-conditioned Invertible Networks (PINs) placed around a differentiable LBS block. These PINs ($\bH_c$ and $\bH_d$) capture non-linear surface deformations of clothes and attenuate LBS artefacts. Our canonical representation is not conditioned on the target pose and requires mesh extraction only once. The \textcolor{ForestGreen}{green} shade in Pre-LBS and Deformed spaces indicate the deformations introduced by PINs, and its intensity denotes their magnitudes.} 

    \label{fig:main}
\end{figure*}

\textbf{Searching Canonical Correspondences.} 
While training on raw scans, we are only provided with points in deformed space. To find their possible correspondences in canonical space, we solve for the roots of Equation~\ref{equ:lbs} using an iterative solver while keeping $\bw_{lbs}$ constant. Specifically, we use Broyden's Method~\cite{Broyden1965BOOK} to find a set of $\{\qcpoint^{1}, ..., \qcpoint^{K}\}$ point correspondences for each deformed point $\qdpoint$ by initializing the root-finding algorithm at $K$ different points in the canonical space.  
\begin{align}
\{\qcpoint^{1}, \dots, \qcpoint^{K}\} = \mathbf{broy}(\mathbf{w}_{lbs}, \bpose,  \qdpoint).
\label{equ:broy}
\end{align}

\textbf{Differentiable Skinning.} The above formulation is end-to-end differentiable as it is possible to compute the gradients of the weight field $\mathbf{w}_{lbs}$ with respect to input point $\qdpoint$ via implicit differentiation as shown in SNARF~\cite{snarf}, Section 3.4. 
In this work, we also extend these derivations to compute the gradient of correspondences $\qcpoint^{i}$ w.r.t. input points. For more details, please refer to Appendix~\ref*{a:grad_lbs}.

\textbf{Why is LBS insufficient?} The above differentiable formulation suffers from the same limitations of traditional LBS, such as being unable to represent the clothed surfaces, and introducing volume loss, as shown in Figure~\ref{fig:qual1}, b). This is especially problematic when learning from real-world data of clothed humans in various poses.

\subsection{Pose-conditioned Invertible Network (PIN)}

Invertible networks~\cite{realnvp, nice, kingma2018glow} are bijective functions composed of modular components called coupling layers, which preserve 1-1 correspondences between their input and output. In this section, we describe the construction of our proposed pose-conditioned coupling layer, which is chained together to construct a \pin.

\textbf{2D Coupling Layer (Figure~\ref{fig:coupling}).} A coupling layer operates by splitting its input into two parts using a fixed breaking pattern. After splitting, the first part of the input is transformed by applying a sequence of invertible operations, such as translation and rotation. The parameters for these operations can be produced by any arbitrary function that is jointly conditioned on the second part of the input and an external conditioning, such as pose in our case.  

Formally, as we operate in 3D space, let the input point be defined as $[x,y,z]$, and the input splits are $[x,y]$ and $[z]$. Then the 2D coupling layer $\mathbf{G}^{xy}([x,y,z], \theta^{t})$ defines an invertible transformation as follows: 
\begin{align}
    \label{eq:2dcoupling}
    [x',y'] = \mathbf{R}_{xy} [x,y]^{T} + [{t}_x, {t}_y]~~\text{and}~~z' = z,
\end{align}
where $\mathbf{R}_{xy} \in \mathbb{R}^{2 \times 2}$, and $[t_x, t_y] \in \mathbb{R}^{2}$ is a rotation matrix and translation vector produced by any arbitrary function that takes as input 
only the bone pose $\theta^{t}$ and the coordinate $z$. The inverse $\mathbf{G}^{-1}_{xy}([x,y,z], \theta^{t})$ of the coupling layer can be computed easily by: 
\begin{align}
    [x,y] = \mathbf{R}_{xy}^{-1} ([x',y'] - [{t}_x, {t}_y])~~\text{and}~~z = z'.
\end{align}

We describe computation of operation parameters $\mathbf{R}_{xy}$ and $[t_x, t_y]$ next. 

\textbf{Pose Embedding.} We encode every bone transform in pose $\theta^t$ using a MLP $\mathbf{m}_b$ which takes a 6D input of concatenated 3D translation and rotation (as Euler angles). To obtain pose embedding, we concatenate the outputs of each bone  $\mathbf{e}_{\theta}$ as follows:
\begin{align}
\mathbf{m}_b: \mathbb{R}^{6} \rightarrow \mathbb{R}^{d/n_b}~~\text{and}~~\mathbf{e}_{\theta} :=  \text{\textbf{concat}}[\mathbf{m}_b(\bone_i)]_{i=1}^{n_b}.
\end{align}

\textbf{Space Embedding.} We use SIREN~\cite{sitzmann2019siren}, a learned and periodic positional encoding, to map the spatial coordinates denoting it as 
\begin{equation}
    \label{eq:siren}
    \mathbf{e}_{z} := \mathbf{\Phi}(z): \mathbb{R}^{1} \rightarrow \mathbb{R}^{d}.
\end{equation}
We find that this helps to better represent high-frequency surface details such as cloth wrinkles. 

\textbf{Space and Pose Aware Conditioning (Figure~\ref{fig:sp_cond}).} \textit{We observe that when the relative pose $\btheta^t$ between deformed and canonical spaces is zero} (\ie $\bone_i = [\mathbf{I} | \mathbf{0}]$, all bone transforms have identity rotation and zero translation), \textit{the coupling layer should not introduce any space-varying (\ie $z$-conditioned) changes}. 

To enforce this, we take the Hadamard product of the space and pose embeddings, and subsequently concatenate them obtaining 
\begin{align}
    \mathbf{e}_{sp} := \text{\textbf{concat}}[\mathbf{e}_{\theta} \odot \mathbf{e}_{z}, \mathbf{e}_{\theta}] \in \mathbb{R}^{2d}.
\end{align}

We visualize the construction of our space and pose aware conditioning  $\mathbf{e}_{sp}$ in Figure~\ref{fig:sp_cond}. 

\textbf{Rotation and Translation Maps.} Finally, to produce parameters for coupling operations, we use two MLPs $\mathbf{m}_t$ and $\mathbf{m}_r$, which take as input the above conditioning vector: 
\begin{align}
    \label{eq:2dmaps}
    [{t}_x, {t}_y] = \mathbf{m}_{t}(\mathbf{e}_{sp}): \mathbb{R}^{2d} \rightarrow \mathbb{R}^{2}, \\ 
    \gamma_{xy} = \mathbf{m}_{r}(\mathbf{e}_{sp}): \mathbb{R}^{2d} \rightarrow \mathbb{R}^{1}.
\end{align}
Note that the output of $\mathbf{m}_{r}$ only predicts the angle of rotation $\gamma_{xy}$ in radians (a single value). The axis of rotation passes through the origin of the split input space, \ie $\mathbf{XY}$ space here. We convert $\gamma_{xy}$ into a rotation matrix $\mathbf{R}_{xy}$.

\textbf{1D Coupling Layer.} Unlike the 2D coupling layer described above, we cannot use the rotation operator in 1D, and in this case, we only use translation. For layer $\mathbf{G}^{x}([x,y,z], \theta^{t})$ with split pattern as $[x]$ and $[y,z]$ the coupling operation becomes: 
\begin{align}
    x' = x + {t}_x~~\text{and}~~[y', z'] = [y,z],
\end{align}
where $t_x \in \mathbb{R}^{1}$ produced in a similar fashion as a 2D coupling layer using a translation map $\mathbf{m}_t : \mathbb{R}^{2d} \rightarrow \mathbb{R}^{1}$ with single scalar output instead of 2D translation. The space embedding of Equation~\ref{eq:siren}, in the 1D case, takes both coordinates as input $ \mathbf{e}_{xy} = \bPhi([x,y]): \mathbb{R}^{2} \rightarrow \mathbb{R}^{d}$.        

\begin{table*}[!t]
\centering
\resizebox{\textwidth}{!}{
\begin{tabular}{ll ccccc ccccc}
\toprule
{} & {} &\multicolumn{5}{c}{\textbf{IoU Surface}} & \multicolumn{5}{c}{\textbf{IoU Bounding Box}}\\
\cmidrule(r){3-7} \cmidrule(r){8-12}
\textbf{Subject} & \textbf{Clothing} & \textbf{AVG-LBS} & \textbf{FIRST-LBS} & \textbf{SNARF} & \textbf{SNARF-NC} & \textbf{INS (ours)} & \textbf{AVG-LBS} & \textbf{FIRST-LBS} & \textbf{SNARF} & \textbf{SNARF-NC} & \textbf{INS (ours)}\\
\midrule

\textbf{Average} &  & 65.01\% & 57.41\% & \underline{72.24\%} & 66.89\% & \textbf{73.13\%} & 65.12\% & 57.5\% & \underline{72.17\%} & 66.78\% & \textbf{73.19\%}  \\

\bottomrule
\end{tabular}
}
\caption{\textbf{Quantitative Results on Clothed Humans.} We find our approach INS outperforms all methods when averaged across 15 runs, on both IoU Surface and IoU Bounding Box metrics.  \vspace{-1em}}
\label{tab:cape}
\end{table*}

\textbf{Pose-conditioned Invertible Network (\pin).} Finally, we compose our PIN by chaining together multiple 1D and 2D pose-conditioned coupling layers as:
\begin{align}
    \label{eq:compose_pin}
    \mathbf{H}(\mathbf{p}, \boldsymbol{\theta}^{t}) = \mathbf{G}_{1} \circ \mathbf{G}_{2} \circ \dots \mathbf{G}_{n} : \mathbb{R}^{3} \rightarrow \mathbb{R}^{3},
\end{align}
where $\mathbf{p}$ represents point in 3D space, $\bG_i$ represents a coupling layer, and $\boldsymbol{\theta}^{t}$ represents pose. Inverting PIN is simply equivalent to sequentially inverting each coupling layer in the reverse order:
\begin{align}
    \mathbf{H}^{-1}(\mathbf{p}, \boldsymbol{\theta}^{t}) = \mathbf{G}_{n}^{-1} \dots \mathbf{G}_{2}^{-1} \circ \mathbf{G}_{1}^{-1}.
\end{align}
Since the PIN is invertible by construction, it preserves exact correspondences between its input and output spaces:
\begin{align}
    \mathbf{p} = \mathbf{H}^{-1}(\mathbf{H}(\mathbf{p}, \boldsymbol{\theta}^{t}), \boldsymbol{\theta}^{t})~~\forall~~\mathbf{p} \in \mathbb{R}^3.
\end{align}

We visualize a single coupling layer of \pin  in Figure~\ref{fig:coupling}. 

\subsection{Invertible Neural Skinning}
\label{ssec:ins}
\textbf{Overview (Figure~\ref{fig:main}).} Our overall posing pipeline INS is comprised of three previously described components chained together: 
\begin{compactitem}
    \item $\bH_c$: A Pose-conditioned Invertible Network (PIN) $\bH_c$ that operates after canonical space and before LBS. 
    \item $\bH_d$: A Pose-conditioned Invertible Network (PIN) that operates before deformed space and after LBS.
    \item Differentiable LBS network as described in Section~\ref{ssec:difflbs} operating between above PINs.
\end{compactitem}
Next, we discuss how we formulate an invertible mapping that preserves correspondences between deformed and canonical spaces.

\textbf{Deformed to Canonical (Training).} For any point $\dpoint$ in deformed space, we first process it using PIN $\mathbf{H}_{d}$, and obtain $\mathbf{q}^t_{d}$. Next, we use Broyden's algorithm to get correspondences of $\mathbf{q}^t_{d}$ in canonical space, let's say $\{\mathbf{q}^i_{c}\}_{i=1}^K$. Finally we use a second PIN $\mathbf{H}_{c}$ to map these points $\{\mathbf{p}_{c}\}_{i=1}^K$ in the pose-independent canonical space. 
\begin{align}
\dpoint \xrightarrow[]{\mathbf{H}_{d}(.,\boldsymbol{\theta}^{t})} \qdpoint \xrightarrow[]{\mathbf{broy}(.,\bpose)} \{\qcpoint^{i}\} \xrightarrow[]{\mathbf{H}_{c}(.,\boldsymbol{\theta}^{t})} \{\cpoint^{i}\}.
\end{align}
To obtain the most suitable canonical correspondence, we take the $\argmax$ over all predicted canonical occupancies
\begin{align}
\cpoint^* = \argmax_{i=1...K} \{  \mathbf{O}(\cpoint^i) \}.
\label{equ:compose}
\end{align}
During training, we approximate the \textbf{$\argmax$} with a $\mathrm{softmax}$ function in order to backpropagate gradients softly through all correspondences following SNARF. 

\textbf{Training Objective.} In our dataset, we are given points in deformed space and corresponding ground truth occupancy values of zero or one. We map these deformed points to canonical space and apply binary cross-entropy loss to jointly train all components of the posing network according to 
\begin{align}
    \min_{\innd, \innc, \mathbf{w}_{lbs}, \mathbf{O}}\mathcal{L}_{bce}(\mathbf{O}(\cpoint), o_{gt}).
    \label{eq:bce}
\end{align}

\textbf{Auxiliary Objectives.} Following SNARF, we enforce a prior on the canonical pose by using two additional losses during the first epoch. First, we sample additional points on bones in canonical pose and encourage their occupancies to be one. Second, we encourage the skinning weight of bone joints to be equal. However, no ground truth skinning weights are required during these steps.   

\textbf{Canonical to Deformed (Inference).} Once trained, we can animate characters using INS in any given novel pose $\boldsymbol{\theta}^n$ in two simple steps. First, running mesh extraction on the canonical occupancy network. Second, reposing the mesh vertices via an inverse pass of our posing pipeline as follow:     
\begin{align}
\label{eq:inf}
\cpoint \xrightarrow[]{\mathbf{H}^{-1}_{c}(.,\pose^{n})} \qcpoint \xrightarrow[]{\mathbf{lbs}(.,\pose^{n})} \qdpoint \xrightarrow[]{\mathbf{H}^{-1}_{d}(.,\pose^{n})} \dpoint.
\end{align}

\textbf{Fast Reposing.} As our canonical occupancy network $\bO$ is independent of $\theta^n$ we only have to extract mesh exactly once. And reposing this mesh for a sequence of poses simply becomes equivalent to performing multiple inferences described in Equation~\ref{eq:inf}.  

\section{Experiments}
\begin{table}[!t]
\centering
\resizebox{\columnwidth}{!}{
\begin{tabular}{l ccc ccc}
\toprule
{} &\multicolumn{3}{c}{\textbf{IoU Surface}} & \multicolumn{3}{c}{\textbf{IoU Bounding Box}}\\
\cmidrule(r){2-4} \cmidrule(r){5-7}
\textbf{Subject} & \textbf{SNARF} & \textbf{SNARF-NC} & \textbf{INS (ours)} & \textbf{SNARF} & \textbf{SNARF-NC} & \textbf{INS (ours)}\\
\midrule

\textbf{Average} & \textbf{90.01\%} & 85.22\% & \underline{88.59\%} & \textbf{97.21\%} & 95.72\% & \underline{96.35\%}  \\

\bottomrule
\end{tabular}
}
\caption{\textbf{Quantitative Results on Minimally Clothed Humans.} On DFAUST, INS outperforms SNARF-NC by a large margin while performing competitively with SNARF, and being order magnitude faster at reposing.     
}
\label{tab:dfaust}
\vspace{-0.5cm}
\end{table}

\subsection{Evaluation}

\textbf{Datasets.} 
Training our method requires sampled points in the deformed space, along with corresponding occupancies and poses. Thus, we benchmark INS on two datasets CAPE~\cite{CAPE:CVPR:20}, which features scanned humans in loose clothing, and DFAUST~\cite{AMASS:ICCV:2019}, containing only minimally clothed human scans. 

CAPE~\cite{CAPE:CVPR:20}  contains scans of 15 subjects (8 males and 7 females), wearing 8 different types of garments while performing a large number of actions. These actions were recorded using a high-resolution body scanner (3dMD LLC, Atlanta, GA), and the scans were registered using an SMPL model~\cite{SMPL:2015}. Similarly to SNARF~\cite{snarf}, INS requires training a new model every \textit{subject-cloth} pair, and exhaustively training on every combination quickly becomes expensive. To manage computational costs, we use a subset of 15 sequences. This subset covers all garment types at least once and most of the subjects, thus capturing variations in both --- body shape and clothing. 

DFAUST is a subset of the AMASS~\cite{AMASS:ICCV:2019} dataset consisting of 10 subjects who are minimally clothed. Each subject is scanned similarly to CAPE while performing 10 different actions. As the subjects wear minimal clothing, much of their motions can be represented accurately by rigid body transformations. As a consequence of little subject clothing, we observe that DFAUST contains significantly fewer pose-specific deformations. Thus we note that DFAUST dataset is not well-posed to test the true capabilities of INS.

\textbf{Data Splits.} For a given subject in DFAUST or a subject-clothing pair in CAPE, we are provided with multiple temporal sequences, each containing a different action. We divide these sequences in 9:1 ratio into train and test sets. This split is similar to SNARF~\cite{snarf}. More details about training sequences and garment types are provided in Appendix~\ref*{a:data}.

\textbf{Metrics.} Following SNARF~\cite{snarf}, we report the mean Intersection-over-Union of points sampled near the mesh surface (IoU surface), and of points sampled uniformly in space (IoU bbox). 
\begin{figure}
    \centering
    \includegraphics[width=0.45\textwidth]{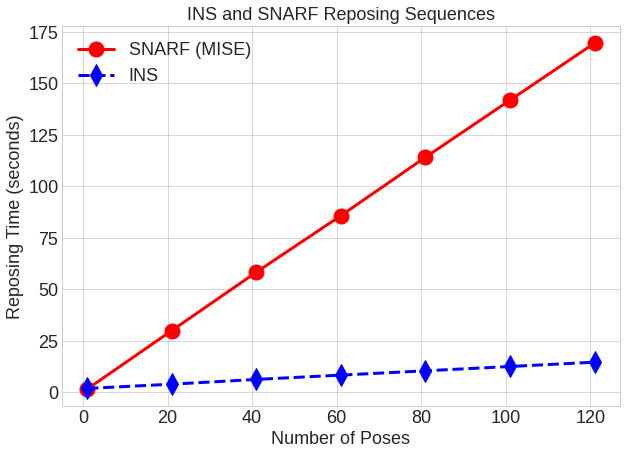}
    \captionof{figure}{\textbf{Reposing time comparison between INS and SNARF} We show the time taken by SNARF vs INS for reposing a mesh extracted at $128^3$ resolution across 125 different target poses. INS performs reposing an order of magnitude faster than SNARF.}
    \label{plot:time}
    \vspace{-0.3cm}
\end{figure}

\subsection{Baselines}

\textbf{SNARF-NC.} We use SNARF~\cite{snarf} without pose-conditioning in the canonical occupancy network as our first and primary baseline. For this, we only remove the pose-conditioning used by SNARF such that canonical space becomes pose-independent, \ie $\mathbf{O(\cpoint)}$. We do not make any other changes. This setting is comparable to INS as it allows for fast posing and preserves correspondences across different poses. 

\textbf{SNARF.} We also compare INS to the original SNARF~\cite{snarf} which uses a pose-conditioned occupancy network, \ie $\mathbf{O(\cpoint, \mathbf{\theta}^t)}$. However, we point out that the above pose-conditioned occupancy comes at the sacrifice of fast posing, by requiring expensive mesh extraction for each new pose while not preserving correspondences across them. These disadvantages make the direct comparison between INS and SNARF based solely on their performance a little lopsided.    

To obtain SNARF and SNARF-NC results, we use the official codebase~\footnote{\url{https://github.com/xuchen-ethz/snarf}} released by the authors. 

\textbf{AVG-LBS.} In addition to the above strong learned baselines, we provide results on two simpler baselines, which use the SMPL-fitted LBS weights to unpose the meshes (scans) using forward skinning. For this, we simply take an average of all the canonicalized training meshes to generate a final canonical mesh and deform it to any unseen given pose using Foward LBS and SMPL weights.       

\textbf{FIRST-LBS.} This baseline is similar to the AVG-LBS baseline described above and uses SMPL-fitted weight for reposing. Instead of using an average across all training meshes, it only uses the first mesh, thus containing lesser pose-conditioned details. 
\begin{figure}
    \centering
    \includegraphics[width=0.5\textwidth]{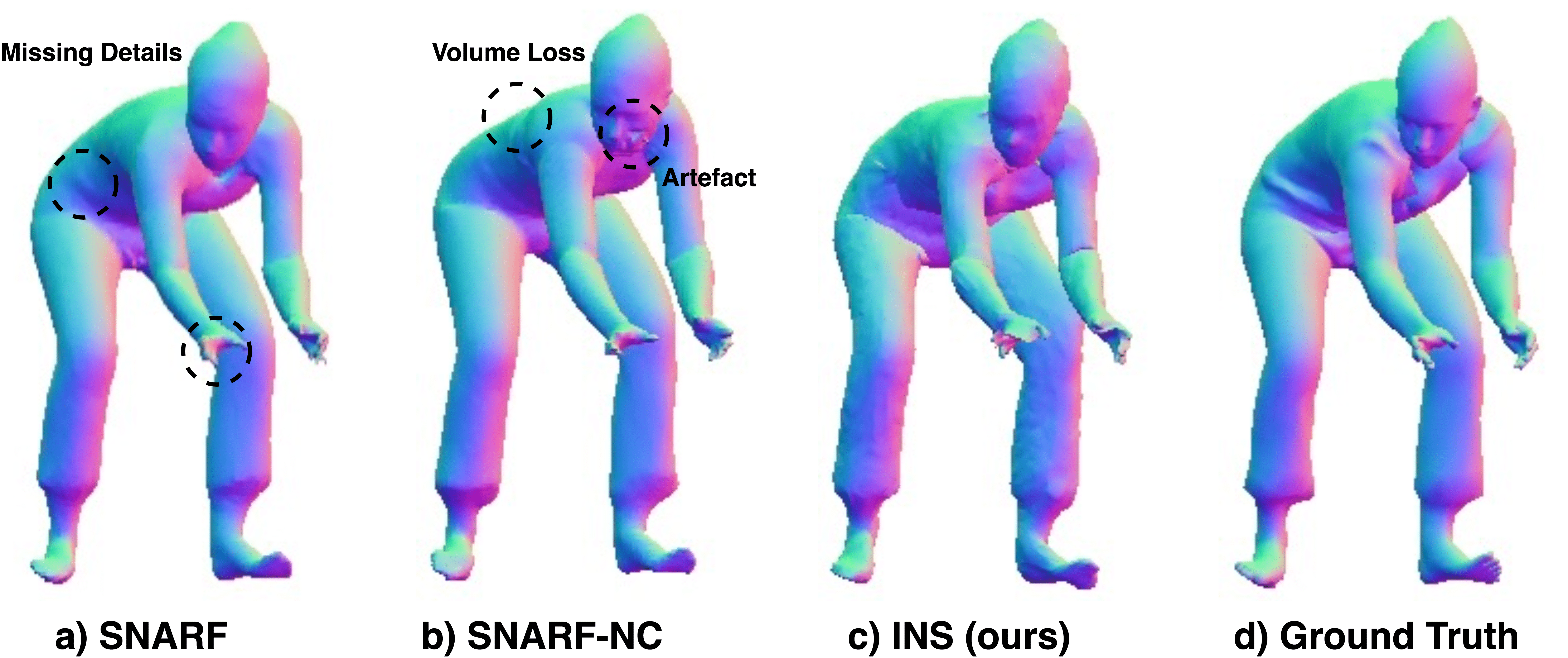}
    \captionof{figure}{\textbf{Qualitative Samples on CAPE.}  We find that SNARF (leftmost) struggles to represent finer details such as cloth wrinkles. Whereas SNARF-NC (second left) struggles with LBS artefacts such as volume loss, and candy-wrapper effects. Meanwhile INS (second right) is able to repose surfaces while capturing sharper local-details. Best viewed under zoom.}
    \label{fig:qual1}
    \vspace{-0.3cm}
\end{figure}

\subsection{Main Results}

\textbf{CAPE.} We demonstrate the results of INS on clothed human data in Table~\ref{tab:cape}. Given the challenging nature of modeling cloth deformations contained in this dataset, we find that INS surpasses SNARF-NC (without pose-conditioning) on average by \textbf{+6.24\%} and \textbf{+6.41\%} absolute percentage points in Surface IoU and Bounding Box IoU respectively. Moreover, INS also outperforms vanilla SNARF with pose-conditioning by \textbf{+0.89\%} and \textbf{+1.02\%} absolute percentage points in Surface IoU and Bounding Box IoU, respectively, while also enjoying the benefits of fast posing, and matched correspondences across various poses. We observe that the simple aggregation baseline of AVG-LBS performs quite closely with SNARF-NC with a performance drop of only \textit{1.88\%} and \textit{1.66\%} percentage points between them. However, AVG-LBS benefits from using a strong prior of parametric SMPL model and corresponding fitted weights.

\textbf{DFAUST.} We also report results on much simpler minimally clothed humans from the DFAUST dataset in Table~\ref{tab:dfaust}. We find that INS outperforms SNARF-NC (without pose-conditioning) on average by \textbf{+3.37\%} and \textbf{+0.63\%} absolute percentage points in Surface IoU and Bounding Box IoU metrics, respectively. When compared to SNARF with pose conditioning, we find INS lags behind by \textbf{-1.42\%} and \textbf{-0.86\%} absolute percentage points in Surface IoU and Bounding Box IoU metrics, respectively. Given the minimal clothing and few pose-conditioned non-linear effects in DFAUST, we hypothesize that this performance drop can be attributed to SNARF overfitting easily to this benchmark. We believe this result also reflects the importance of testing on many realistic datasets such as CAPE.  

\textbf{Timing Study.} In Figure~\ref{plot:time}, we compare the times taken by SNARF and INS to repose a clothed character across a sequence of 125 different poses. A single mesh extraction pass with MISE~\cite{mescheder2019occupancy} operating on the cube of resolution $128^3$, takes nearly $1.5$ seconds. While reposing, SNARF performs this operation for every given pose, whereas INS requires mesh extraction only once. \textit{Reposing the extracted mesh INS takes $0.13$ seconds for an inference pass, which is an order of magnitude faster than SNARF.}   

\begin{table}[!t]
\centering
\resizebox{\columnwidth}{!}{
\begin{tabular}{llcc}
\toprule
$\var{\#}$ & \textbf{Ablation} & \textbf{IoU Surface ($\%$)} & \textbf{IoU Bounding Box($\%$)}\\
\midrule

$\var{1}$ 
 & \ins (vanilla) 
 & \textbf{72.83} 
 & \textbf{72.69} \\
$\var{2}$ 
 & w/o Pose Mul. 
 & $61.94_{\color{red}-\mathbf{10.89}}$ 
 & $62.00_{\color{red}-\mathbf{10.69}}$  \\
$\var{3}$ 
 & w/o SIREN 
 & $69.67_{\color{red}-3.16}$ 
 & $69.57_{\color{red}-3.12}$  \\
$\var{4}$ 
 & w/o Rotation 
 & $71.91_{\color{red}-0.92}$ 
 & $71.87_{\color{red}-0.82}$  \\
$\var{5}$ 
 & w/o $\mathbf{H}_{d}$ 
 & $72.66_{\color{red}-0.17}$ 
 & $72.58_{\color{red}-0.11}$  \\
$\var{6}$ 
 &  w/o $\mathbf{H}_{c}$ 
 & $67.89_{\color{red}-\underline{4.94}}$ 
 & $67.81_{\color{red}-\underline{4.88}}$ \\
$\var{7}$ 
 &  w/o \textbf{LBS} 
 & ${40.79}_{\color{red}-32.04}$ 
 & ${40.65}_{\color{red}-32.04}$ \\
\bottomrule
\end{tabular}
}
\caption{\textbf{Ablation Table.} We perform an ablation study of INS on a clothed subject 03375 (Table~\ref{tab:cape}, Row 1) from the CAPE dataset.}
\label{tab:ablation}
\vspace{-0.5cm}
\end{table}

\subsection{Ablations}

We perform numerous ablations of our INS setup, the results of which are summarized in Table~\ref{tab:ablation}. 

\textbf{Multiplying Pose and Space Embeddings is important.} Reformulating the pose conditioning by simple concatenation, \ie $[\mathbf{e}_{z}, \mathbf{e}_{\theta}]$ instead of multiply-then-concatenate, \ie $[\mathbf{e}_{z}, \mathbf{e}_{z} \odot \mathbf{e}_{\theta}]$ leads to a significant performance drop of $\sim$ 11\% points in both metrics (Row 2).

\textbf{$\mathbf{H}_{c}$ contributes much more  than $\mathbf{H}_{d}$.} The invertible networks do not contribute equally to the performance. Removing the canonical space PIN $\mathbf{H}_{c}$ leads to a sharp drop in IoU Surface by 4.94\%, when compared to removing the deformed space INN  $\mathbf{H}_{d}$ which drops performance by only 0.17\% points (Rows 5,6). We attribute this partly to the fact that editing an LBS-deformed mesh introduces the additional complexity of resolving part-wise correspondences, such as locating the new positions of joints and limbs.

\begin{figure*}[t!]
    \centering    
    \includegraphics[width=\linewidth]{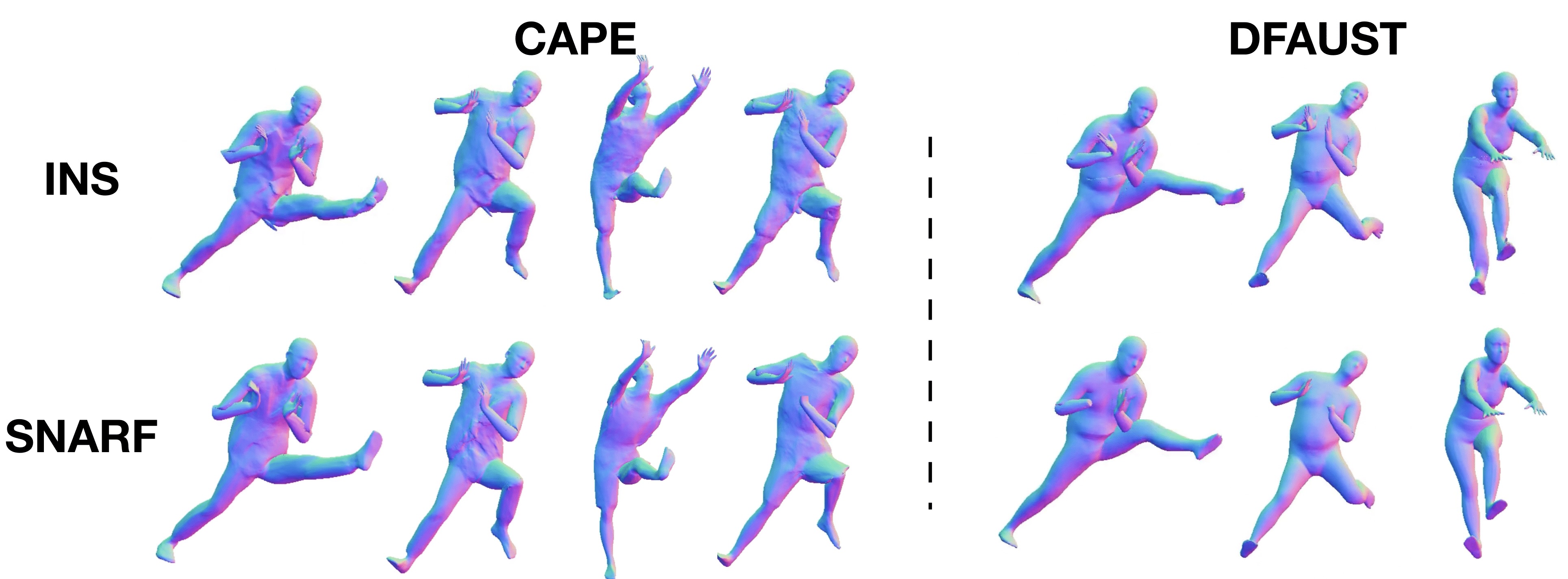}
    \captionof{figure}{\textbf{Comparison of INS and SNARF models undergoing extreme pose animation from PosePrior dataset.}}
    \label{fig:pp}
\end{figure*}

\textbf{Replacing SIREN positional embedding with MLP hurts.} IoU Surface drops by 3.16\% if the learned sinusoidal embeddings are replaced by simple MLP layers. This happens as fine surface details such as cloth wrinkles get blurred when using MLP, which is prevented by SIREN.     

\textbf{PINs without rotation perform slightly worse.} Removing 2D operations from our PINs leads to a drop in IoU Surface by 0.92\%. This happens because twisting deformations in the surface have to be represented by only displacements, which has previously been shown to be difficult to learn~\cite{nerfies}.

\textbf{Simply using PINs without LBS performs worse.} Entirely removing the differential LBS module and relying solely on PINs to capture the full articulate motion results in a huge drop of $32$\% on both metrics (Row 7).

\subsection{Qualitative Analysis}

\textbf{INS can represent finer details compared to SNARF.} In Figure~\ref{fig:qual1}, we demonstrate a subject in a challenging novel pose from the CAPE dataset. We find that SNARF (leftmost) is unable to capture fine details of cloth wrinkles, while also missing fingers as highlighted by the markers. Whereas SNARF-NC (second left) struggles with LBS artefacts such as volume loss by shrinking the arched back (highlighted), and displaying candy-wrapper effects. Finally, INS (second right) is able to capture much sharper local details around the body joints, such as around the waist and neck.   

\begin{figure}[t!]
    \centering
    \includegraphics[width=0.5\textwidth]{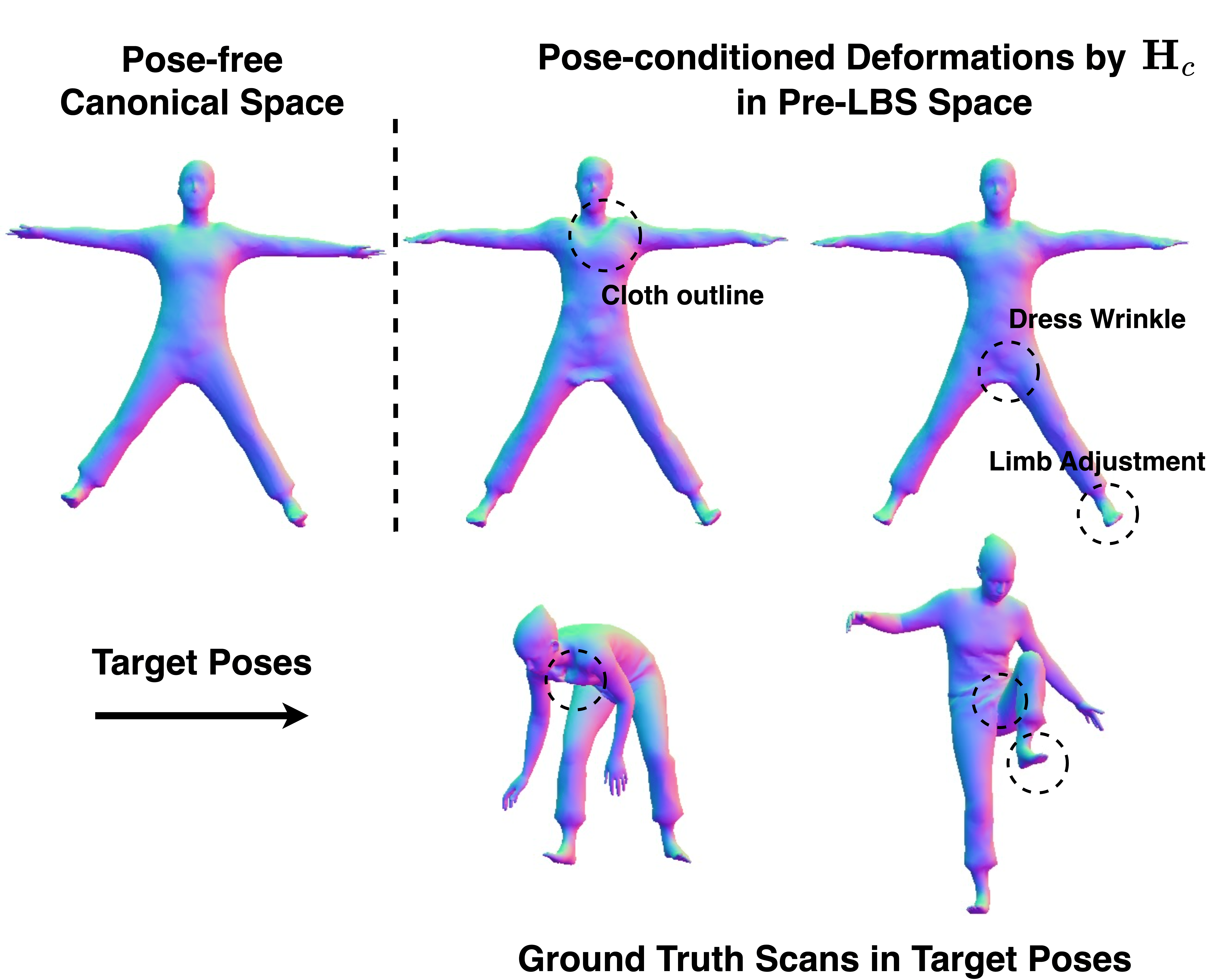}
    \captionof{figure}{\textbf{Pre-LBS space deformations.} We show the edits made by PIN $\bH_c$ before LBS operation for two novel target poses. Best viewed under zoom.}
    \label{fig:qual2}
    \vspace{-0.5cm}
\end{figure}

\textbf{PINs can represent pose-varying deformations well.} In Figure~\ref{fig:qual2}, we tease apart the edits made by solely PIN $\bH_c$ in the canonical space (displayed in the top row) given two unseen target poses (shown in the bottom row). As highlighted in the figure, we find that PIN learns to introduce pose-varying deformations such as raising cloth outlines around the neck and shoulder joint, introducing dress wrinkles at near extremities, and even adjusting limbs such as orienting feet. 

\textbf{Extreme poses from PosePrior.} The most loose-fitted sequence in CAPE is of subject \texttt{00375} wearing a blazer and trouser, we show it animated in extreme poses from Pose Prior in first two columns of Figure~\ref{fig:pp}. We also visualize other subjects (both clothed and naked) in extreme poses as well. We find that INS produces much more realistic cloth deformations compared to SNARF. 

\section{Conclusion}
In this work, we presented an invertible, end-to-end differentiable, and trainable pipeline called Invertible Neural Skinning for reposing humans. For this, we built a Pose-conditioned Invertible Network (PIN) that can handle non-linear surface deformations of clothes and skin well, while also retaining correspondences across different poses. By placing two PINs around a differentiable LBS network and using a pose-free occupancy network, we created INS. We show that INS outperforms previous methods on clothed humans, while staying competitive on simpler and minimally clothed humans. Since reposing with our method requires the expensive mesh extraction exactly once, INS provides a speed-up of an order of magnitude compared to previous methods when animating long pose sequences.     

\textbf{Future Work.} While training INS, the correspondence search performed by the differentiable LBS module often becomes a bottleneck, and future works can explore the possibility of eliminating this module completely --- by learning its behavior from data. Furthermore, the occupancy network can be replaced with a neural representation that can handle texture and lightning and thus learn directly from 2D images and videos instead of raw scans.

\appendix

\begin{table*}[!t]
\centering
\resizebox{0.8\textwidth}{!}{
\begin{tabular}{clccclc}
\toprule
\# & Hyperparameters & Value & & \# & Hyperparameters & Value\\
\midrule
\small\texttt{1} & No. of Parameters in INS & 1.80M &&
\small\texttt{2} & No. of Coupling Layers in $\bH_d$/$\bH_c$ & 18 \\
\small\texttt{3} & No. of Parameters in PIN $\bH_d$/$\bH_c$ & 0.41M &&
\small\texttt{4} & No. of Parameters in Occupancy Network $\bO$ & 0.46M \\
\small\texttt{5} & No. of Parameters in LBS Network & 53K &&
\small\texttt{6} & No. of Parameters in Bone Encoder & 0.46M \\
\small\texttt{7} & Pose Embedding Dimension & 120  &&
\small\texttt{8} & Space Embedding Dimension & 120  \\
\small\texttt{9} & Space and Pose Embedding & 240 &&
\small\texttt{10} & PIN input and output Dimension & 3 \\
\small\texttt{11} & Number of epochs & 250 &&
\small\texttt{12} & Optimizer & Adam \\
\small\texttt{13} & Batch size (DFAUST/CAPE) & 12/8 &&
\small\texttt{14} & Learning rate (INNs/Rest) & 1e-4/1e-3 \\
\small\texttt{15} & Warm-up learning rate factor & 0.2 &&
\small\texttt{16} & Warm-up iterations & 2400 \\
\small\texttt{17} & No. of points per batch & 60000 &&
\small\texttt{18} & Gradient clipping (L-2 Norm) & 4.0 \\
\midrule
\end{tabular}
}
\caption{Hyperparameters and Training configuration to train INS.} 
\smallskip
\label{tab:hyperparameters_supp}
\end{table*}

\section{Implementation Details}
\subsection{Sampling Points}
\label{a:sampling}
Following SNARF, we sample 200K points at every frame of the sequence. Half of these points (100K) are near the mesh (scan) surface, which are obtained by first sampling points on the mesh surface via Poisson disk sampling and followed by displacement with isotropic Gaussian noise (of $\sigma = 0.01$). Remaining half (100K) points are sampled uniformly within a bounding box scaled to 110\% of the original bounding box. 

\subsection{Hyperparameters and Training Details}
\label{a:hyperparam}
We trained all our models on a single Tesla V100 GPU for 250 epochs, which took nearly 40 hours on average. We used a learning rate of $1e{-4}$ to train the PINs, while using a learning rate of $1e{-3}$ for remaining modules. We used Adam~\cite{kingma2014adam} optimizer, with a linear warmup and no learning rate decay.  PyTorch~\cite{Paszke_PyTorch_An_Imperative_2019} is used for all the experiments. Please refer to Table~\ref{tab:hyperparameters_supp} for full list.

\subsection{Metrics}
\label{a:metrics}
Given set of sampled points $\bP$ to be evaluated, we can represent the joint tuple of any point, its ground truth occupancy (which can be either 0 or 1), and predicted occupancy as ($\bp_d^i, g^{i}, h^{i}) ~~\forall~~\bp_d^i \in \bP $ respectively. Then Intersection over Union (IoU) can be computed as follows:
\begin{align}
    \text{IoU} =  \sum_{\substack{\bp_d^i \in \bP}} \frac{g^i \cap h^i}{g^i \cup h^i}
\end{align}
To convert predicted probability to binary occupancy, we simply check if it is greater than 0.5. IoU Bounding Box operates with points sampled uniformly in the space, whereas Surface IoU operates with points sampled close to the body as described in Section~\ref{a:sampling}.  

\section{Data}
\label{a:data}
\textbf{CAPE.} CAPE originally contains 15 subjects, with each subject wearing 1-6 different types of clothing, and performing 3-74 different actions. On average, it contains nearly 249 frames for every \textit{subject-cloth} pair. Due to high variance as well as high number of \textit{subject-cloth} pairs, we use a subset of CAPE which contains 15 sequences of 13 subjects containing all  8 different types of clothings. A clothing in CAPE is denoted by a joint string of upper and lower body garment, for example, a subject wearing a blazer and pants is annotated as \textit{blazerlong}, and so on.

\section{Invertible Neural Network}

\subsection{Initialization}
We found that initializing the Pose-conditioned Invertible Networks (PINs) as identity modules stabilizes training, and allows the LBS network to train better. For this, we initialize the weights and biases of the last layer of the operation maps $\bm_r$ and $\bm_t$ (shown in Figure~2) as zeros.

\subsection{Volume Preservation}
Previous works in INNs~\cite{realnvp, kingma2018glow, nice} operating on high-dimensional ($\geq$ 512-d) spaces constrained the Jacobian between input and output to an triangular matrix. This ensured that the Jacobian determinant, used for density modeling, was not expensive to compute. Determinant of a triangular matrix is simply multiplication of its diagonal. However, this prevented these works from using 2D operators such as rotation. We note that such a requirement is unnecessary in our setting, where Jacobian determinant is not needed. Additionally, using rotations also helps to preserve volume between the input and output spaces. 

Next, we show that our PINs consisting of only rotations and translations are volume preserving. Note that to show a transform is volume preserving it is sufficient to show that the determinant of the Jacobian of this transform is one. From Equations 5 and 10, we can express the transform represented by a single 2D coupling layer as: 
\begin{gather*} 
    x' = x\mathrm{cos}(\gamma_{xy})  - y\mathrm{sin}(\gamma_{xy}) + t_x \nonumber \\
    y' = x\mathrm{sin}(\gamma_{xy})  + y\mathrm{cos}(\gamma_{xy}) + t_y \nonumber \\
    z' = z 
\end{gather*} 
Then Jacobian of this transform becomes: 
\begin{align}
\bJ = 
\begin{bmatrix}
\mathrm{cos}(\gamma_{xy}) & -\mathrm{sin}(\gamma_{xy}) & 0 \\
\mathrm{sin}(\gamma_{xy}) & \mathrm{cos}(\gamma_{xy}) & 0 \\
0 & 0 & 1 \\
\end{bmatrix}
\label{eq:jac}
\end{align}
And determinant of the above Jacobian is one, \ie $|\bJ|=1$. Since, our PINs are composed of chaining together such coupling layers described in Equation 13, the overall determinant is also one. Hence volume is preserved within PINs.   

\section{Gradients}
\label{a:grad_lbs}
Training INS requires calculating gradients of the Binary Cross Entropoy (BCE) loss $\mathcal{L}_{bce}$ (Equation 18), with respect to all the components. Let the weights of PINs $\bH_c$, $\bH_d$, the LBS network $\bw_{lbs}$, and the Occupancy network $\bO$ be denoted with $\sigma_c$, $\sigma_d$, $\sigma_{lbs}$, and $\sigma_o$ respectively. Backpropagating through the occupancy network $\bO$ and the PIN $\bH_c$ is straightforward:    
\begin{align}
\pardev{\loss_{bce}}{\sigma_o} = \pardev{\loss_{bce}}{\occ} \cdot \pardev{\occ}{\bO(\cpoint)} \cdot \pardev{\bO(\cpoint)}{\sigma_o} 
\label{eq:occ_grad} \\
\pardev{\loss_{bce}}{\sigma_c} = \pardev{\loss_{bce}}{\bO(\cpoint)} \cdot \pardev{\bO(\cpoint)}{\bH_c(\qcpoint^*)} \cdot \pardev{\bH_c(\qcpoint^*)}{\sigma_{c}}
\label{eq:hc_grad}
\end{align}
where $\occ$ is the predicted occupancy. While the gradients for LBS network $\bw_{lbs}$ and second PIN $\bH_d$ are:
\begin{align}
\pardev{\loss_{bce}}{\sigma_{lbs}} = \pardev{\loss_{bce}}{\bH_c(\qcpoint^*)} \cdot \pardev{\bH_c(\qcpoint^*)}{\qcpoint^*} \cdot \pardev{\qcpoint^*}{\sigma_{lbs}} \\
\pardev{\loss_{bce}}{\sigma_{d}} = \pardev{\loss_{bce}}{\qcpoint^*} \cdot \pardev{\qcpoint^*}{\bH_d(\dpoint)} \cdot \pardev{\bH_d(\dpoint)}{\sigma_{d}} 
\end{align}
where $\qcpoint^*$ is the root of the Equation 17, and $\dpoint$ is the input point. Pytorch's automatic differentiation can handle the gradients in Equations 22 and 23. However, to obtain gradients w.r.t. $\qcpoint^*$implicit differentiation is required, similar to SNARF:
\begin{align}
& \mathbf{lbs}(\qcpoint^*, \btheta^t) - \dpoint = \mathbf{0}  \nonumber \\
\Leftrightarrow ~& \pardev{\mathbf{lbs}(\qcpoint^*, \btheta^t)}{\sigma_{lbs}} + \pardev{\mathbf{lbs}(\qcpoint^*, \btheta^t)}{\qcpoint^*} \cdot \pardev{\qcpoint^*}{\sigma_{lbs}} = \mathbf{0} \nonumber \\
\Leftrightarrow ~& \pardev{\qcpoint^*}{\sigma_{lbs}} = - \left(\pardev{\mathbf{lbs}(\qcpoint^*, \btheta^t)}{\qcpoint^*} \right)^{-1}\cdot \pardev{ \mathbf{lbs}(\qcpoint^*, \btheta^t)}{\sigma_{lbs}} 
\label{equ:def_grad}
\end{align}
And we can find gradients of $\qcpoint^*$ with respect to $\qdpoint$ as follows: 
\begin{align}
& \mathbf{lbs}(\qcpoint^*, \btheta^t) - \dpoint = \mathbf{0}  \nonumber \\
\Leftrightarrow ~& \pardev{\mathbf{lbs}(\qcpoint^*, \btheta^t)}{\qcpoint^*} \cdot \pardev{\qcpoint^*}{\bH_d(\dpoint)} + \mathbf{1} = \mathbf{0} \nonumber \\
\Leftrightarrow ~& \pardev{\qcpoint^*}{\bH_d(\dpoint)} = - \left(\pardev{\mathbf{lbs}(\qcpoint^*, \btheta^t)}{\qcpoint^*} \right)^{-1}
\end{align}

\section{Miscellaneous Failed Experiments}
In order to help with a future research in this direction we list several ideas that have been tried in our project, which however did not improve performance.

\subsection{Invertible Residual Layers}
\textbf{Idea.} Beyond the invertible space-splitting layers, we also experimented with using invertible residual layers~\cite{behrmann2019invertible, irevnet}. These layers operate by limiting the Lipschitz constant of the residual branch, which has be less than one in order to guarantee invertibility. Inversion of these layers can be achieved using a fixed-point iteration method, with convergence rate exponential in the number of iterations. 

\textbf{Outcome.} We tried chaining residual layers with coupling layers alternately, and also placing them in the start and end of the invertible networks. However, these setups performed close or worse than without using residual layers, and were slower due to the expensive inversion pass. 

\subsection{Coupling Layers with Scales}
\textbf{Idea.} Previous works CaDeX~\cite{Lei2022CaDeX}, and NeuralParts~\cite{neuralparts} utilized invertible networks with scale and translation operations, following the architecture proposed in RealNVP~\cite{realnvp}. Such a transform can be represented as:
\begin{gather*} 
    x' = x \mathrm{exp}(s_{x})  + t_x \nonumber \\
    y' = y \mathrm{exp}(s_{y})  + + t_y \nonumber \\
    z' = z 
\end{gather*} 
\textbf{Outcome.} When using these layers in our experiments we encountered the following difficulties:
\begin{itemize}
    \item \textbf{Unstable Training.} Floating point overflows occurred frequently during training due to the exponential scaling term. Even after carefully tuned gradient clipping and learning rate schedules, we encountered frequent experiment failures.   
    \item \textbf{Squashing Effect.} Since the scaling operator can lead to very large outputs from INN, generally a sigmoid squashing layer is used at the end to restrict the input to a  fixed range that matches output distribution. However, due to this sigmoid layer, the INN can no longer be initialized as an Identity layer, even when all the rotations are identity and translations are zero. This leads to a squashing artefacts in outputs.
    \item \textbf{Non-volume Preserving.} The Jacobian of these layers~\cite{Lei2022CaDeX} with scaling is not one, previously derived Equation~\ref{eq:jac}. Due to this additional regularization is needed for training. 
\end{itemize}

\subsection{Pose-conditioned 3D rotation and translation layers}
\textbf{Idea.} We tried learning pose-conditioned global 3D rotation and translation layers. We implemented it similar to the coupling layers to predict rotation and translation parameters, but without any space conditioning.  

\textbf{Outcome.} We did not find significant gains using this, and decided against using them in the final version as they had a big memory footprint. These layers often rotate the canonical space creating issues during mesh extraction. 

\section{Visuals (\href{https://www.youtube.com/watch?v=L7MrPzhqPWQ}{video link})}
\textit{We place all the qualitative results in this \href{https://www.youtube.com/watch?v=L7MrPzhqPWQ}{video}, and discuss its contents below.}  
 
\textbf{[Video Part-1] Pose-varying deformations in INS.} In the first part of the video, we visualize the deformations introduced by PINs $\bH_c$ and $\bH_d$ under varying target poses (shown in top right). Deformations introduced by $\bH_c$ are shown in the top-middle part, whereas those introduced by $\bH_d$ are shown in the bottom-left part shaded in green. We demonstrate that INS is able to handle complex deformations of clothing across poses. 
 
\textbf{[Video Part-2] Baseline Comparison.} In the second part of the video, we compare our method INS against all the five baselines discussed in Section 4.2 of the main paper. While both the LBS baselines, and SNARF-NC suffers from artifacts, we see that INS performs much better than other methods.
 
\textbf{[Video Part-3] INS Ablations.} In the third part of the video, we visualize results from various ablations reported in Section 4.4 of the main paper. Here, we find that removing SIREN leads to an overly smooth surface, and removing the LBS network makes it harder for the network to learn limb movements correctly. 
 
\textbf{[Video Part-4] Texture Propagation.}
As INS can preserve correspondences across poses, it becomes possible to propagate mesh attributes such as texture across various time frames. We conducted an experiment to test this, where we applied texture to the pose-independent canonical mesh. Next, we propagated this texture through the INS network. We show the results of this experiment in the fourth (and last) part of the video. We found that the applied texture deformed realistically like clothing, while being consistent across all frames, and was free of jittering.      
 
To contrast and compare with the above experiment, we conducted similar texture propagation using SNARF. Since, SNARF decodes a separate mesh at each time-step, we color this mesh using the same scheme for coloring INS canonical mesh above. Propagating this texture through the LBS block, we find that it frequently leads to jittery artefacts as the texture overflows across semantically different parts. For example, the texture patch E4 applied to the blazer in a particular frame, overflows onto pants in another frame, and so on. 

\subsection{1D and 2D Coupling Layers}
\begin{figure}[!t]
    \centering
    \includegraphics[width=0.45\textwidth]{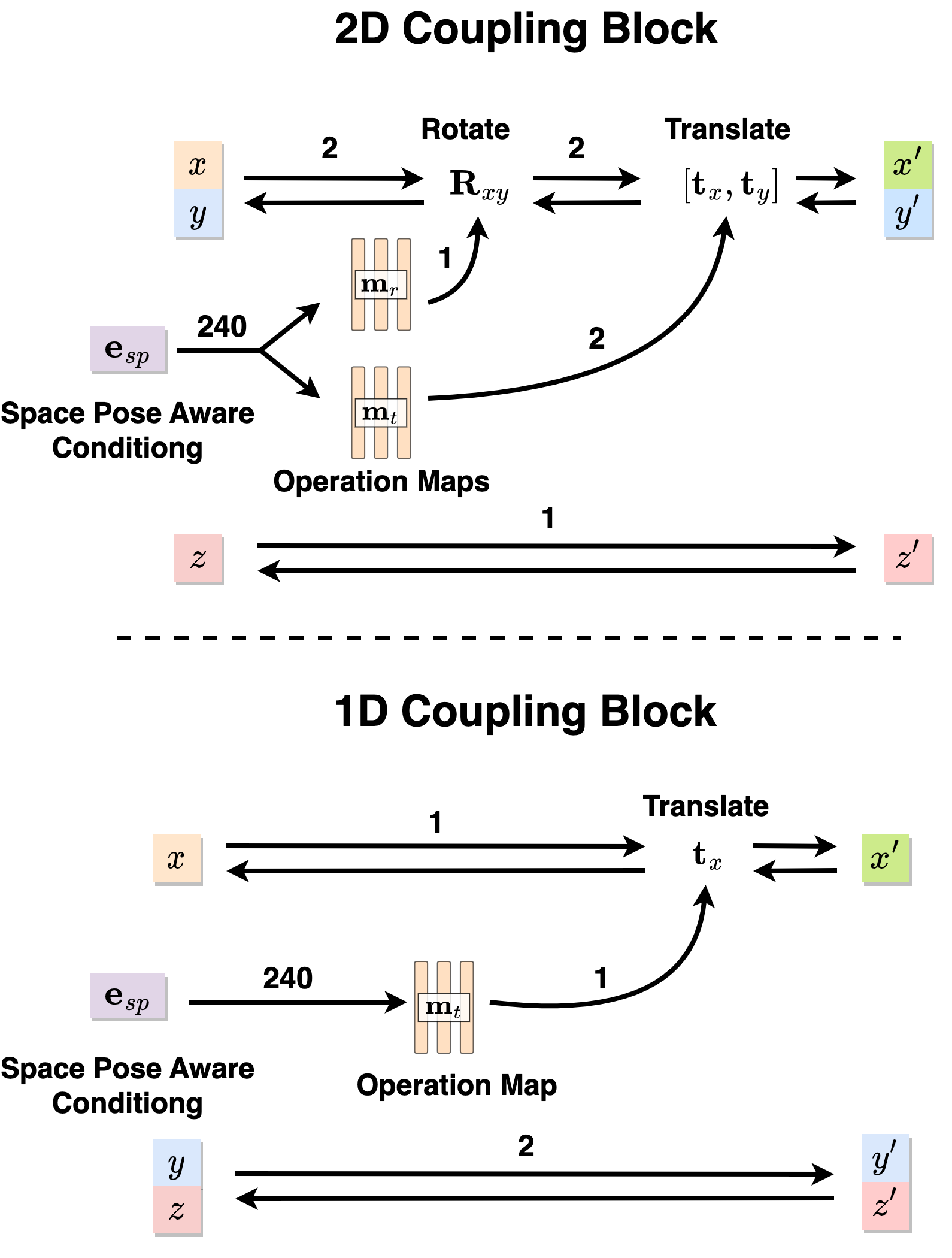}
    \captionof{figure}{\textbf{1D and 2D Coupling Layers.} We show comparison between both types of layers used in PINs. The bidirectional arrows show invertible computations.}
    \label{fig:both-coupling}
    \vspace{-15pt}
\end{figure}

We visualize both 1D and 2D coupling layers together in Figure~\ref{fig:both-coupling} for better understanding. In 1D case the space-pose aware conditioning gets conditioned on the 2D input, which helps to improve expressiveness. In the 2D case, we can make edits on an entire 2D plane conditioned on the 1D input. We find out that these blocks provide complementary benefits, thus we utilize both of them in the final architecture.

\section{Miscellaneous}
\subsection{Training speed of INS comparison.}
Training the LBS module (Broyden's method) takes nearly 0.70s, while both PIN modules combined take 0.09s per iteration. Accounting for loss computation and backprop, single iteration of SNARF takes 1.16s, while INS takes 1.34s. Overall there is a 13.4\% slowdown using INS.

\section{Comparison with IMAvatar}
\noindent \textbf{We can model non-rigid deformations using MLP and leverage Broyden's method in training, similar to IMAvatar~\cite{zheng2022IMavatar}.}

\noindent \textbf{Experiment:} We conducted a prelimnary study, where we used a second MLP to model pose-conditioned vertex offsets before the LBS module similar to IMAvatar. We reused the clothed sequence from ablation study, and name this experiment as \textbf{SNARF-MLP}. \textit{We also added two tricks from INS to make this setup work. First, we zero initialized the last layer of offset MLP; second, we used the space-pose aware conditioning described in Equation 19 (Section 3.2).} 

\begin{table}[!t]
\centering
\renewcommand{\arraystretch}{0.8}%

\resizebox{\columnwidth}{!}{
\begin{tabular}{llccc}
\toprule
$\var{\#}$ & \textbf{Experiment} & \textbf{IoU Surf.} & \textbf{IoU BBox} & \textbf{Train Iter (sec)}\\
\midrule

$\var{1}$ 
 & SNARF-MLP
 & 63.66 
 & 63.74
 & 1.80 \\

$\var{2}$ 
 & w/ Pose Mul. 
 & \underline{68.62} 
 & \underline{68.64} 
 & 1.80 \\

 \midrule
 $\var{3}$ 
 & SNARF-NC
 & 66.10 
 & 66.11
 & \textbf{1.16} \\
 
$\var{4}$ 
 & INS
 & \textbf{72.83} 
 & \textbf{72.69} 
 & \underline{1.34} \\

\bottomrule
\end{tabular}
}
\caption{\textbf{Experiments with setup similar to IMAvatar~\cite{zheng2022IMavatar}.}}
\end{table}

\noindent \textbf{Quantative (Table 1):} 
We found SNARF-MLP performs subpar compared to INS (Row 1 vs 4), while training much slowly due to multiple MLP runs (44\% slowdown). Moreover, we find our pose-conditioning to boosts performance (Row 2), while not using zero init. leads to divergence.
\begin{figure}[t!]
    \includegraphics[width=\linewidth]{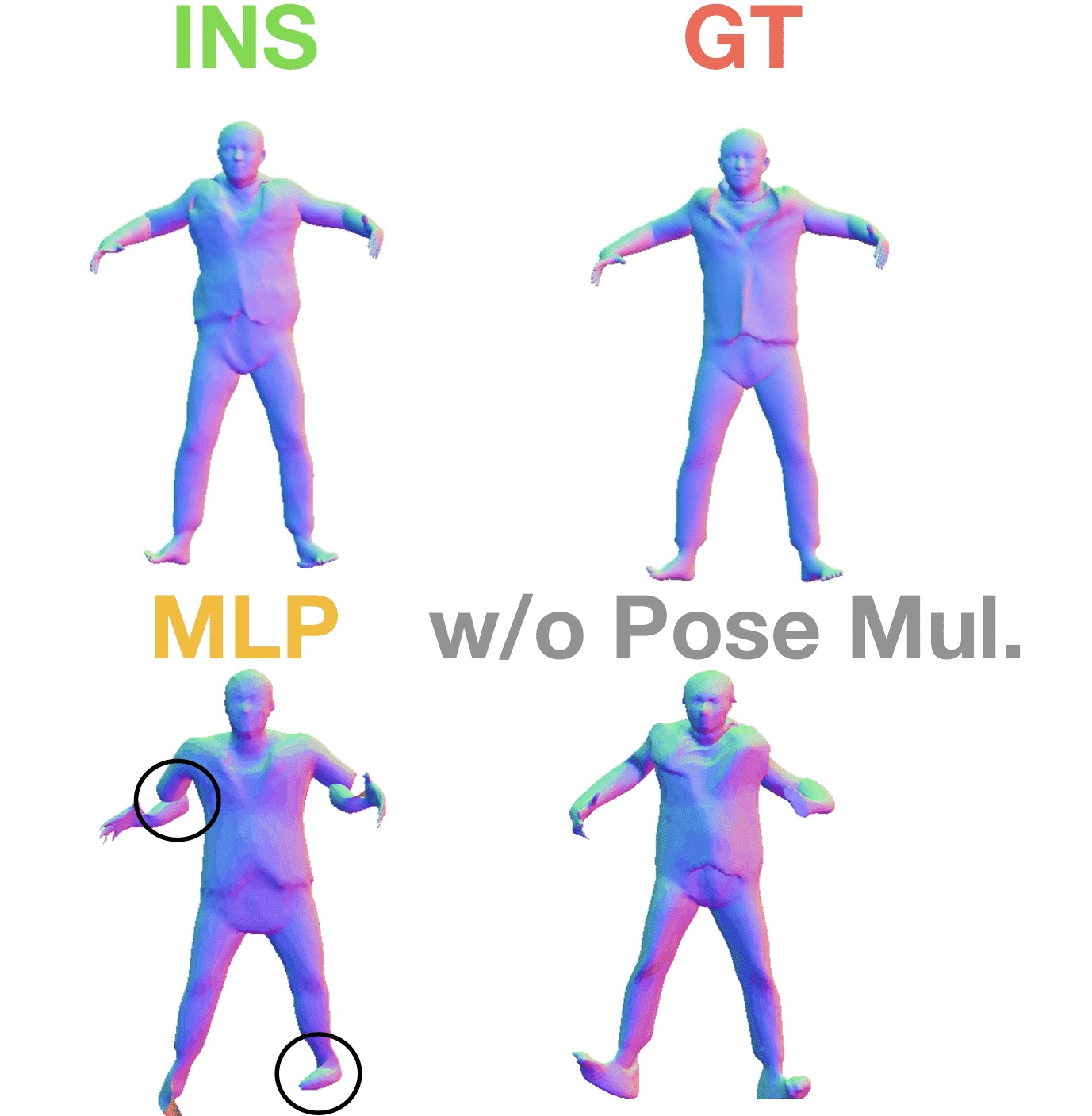}
    \captionof{figure}{\textbf{Visuals showing non-rigid deformations modeled using a second MLP similar to IMAvatar~\cite{zheng2022IMavatar}.}}
    \label{fig:im}
\end{figure}

\noindent \textbf{Qualitative (Fig.1, RHS):} Visually, we found SNARF-MLP to result in unnatural deformations (highlighted). We hypothesize, this could be due to the complex landscape of the resulting function which might make Broyden's vulnerable to local minima. Thus, points close to each other, which ideally should have close solutions, converge to different local minima producing bending artefacts. In contrast, PINs optimized with SGD can avoid local minima~\cite{sgd}.

\begin{table}[!t]
\centering
\resizebox{\columnwidth}{!}{
\begin{tabular}{l ccc ccc}
\toprule
{} &\multicolumn{3}{c}{\textbf{IoU Surface}} & \multicolumn{3}{c}{\textbf{IoU Bounding Box}}\\
\cmidrule(r){2-4} \cmidrule(r){5-7}
\textbf{Subject} & \textbf{SNARF} & \textbf{SNARF-NC} & \textbf{INS (ours)} & \textbf{SNARF} & \textbf{SNARF-NC} & \textbf{INS (ours)}\\
\midrule

03375  & \underline{70.16\%} & 66.1\% & \textbf{71.13\%} & 62.57\% & \underline{70.24\%} & \textbf{71.02\%}  \\

50007 & \textbf{90.28\%} & 83.9\% & \underline{86.63\%} & \textbf{97.77\%} & \underline{96.16\%} & 96.11\%  \\
50022 & \underline{92.19\%} & 88.09\% & \textbf{92.58\%} & \underline{98.05\%} & 96.68\% & \textbf{98.12\%}  \\
50026 & \textbf{91.13\%} & 80.54\% & \underline{89.26\%} & \textbf{97.67\%} & 94.37\% & \underline{97.13\%}  \\
50004 & \textbf{89.6\%} & 85.4\% & \underline{88.96\%} & \textbf{97.48\%} & 96.3\% & \underline{97.21\%}  \\
50009 & \textbf{87.05\%} & 83.87\% & \underline{85.77\%} & \textbf{95.89\%} & \underline{94.63\%} & 93.47\%  \\
50021 & \underline{89.76\%} & 87.26\% & \textbf{90.46\%} & \underline{96.79\%} & 95.58\% & \textbf{96.95\%}  \\
50025 & \underline{90.95\%} & 86.12\% & \textbf{91.59\%} & \underline{97.35\%} & 95.82\% & \textbf{97.55\%}  \\
50027 & \textbf{89.54\%} & \underline{86.9\%} & 85.91\% & \textbf{96.74\%} & \underline{95.79\%} & 93.75\%  \\
50002 & \textbf{89.25\%} & 84.41\% & \underline{85.98\%} & \textbf{97.55\%} & 96.67\% & \underline{97.02\%}  \\
50020 & \textbf{90.31\%} & 85.69\% & \underline{88.74\%} & \textbf{96.85\%} & 95.15\% & \underline{96.23\%}  \\
\midrule
\textbf{Average} & \textbf{90.01\%} & 85.22\% & \underline{88.59\%} & \textbf{97.21\%} & 95.72\% & \underline{96.35\%}  \\

\bottomrule
\end{tabular}
}
\caption{\textbf{Quantitative Results on Minimally Clothed Humans.} On DFAUST, INS outperforms SNARF-NC by a large margin while performing competitively with SNARF, and being order magnitude faster at reposing.     
}
\label{tab:fdfaust}
\vspace{-0.5cm}
\end{table}

\section{Subjectwise result breakdown}
Expanding on the results from main draft, we also report per-subject breakdown on DFAUST (Table~\ref{tab:fdfaust}) and CAPE (Table~\ref{tab:fcape}) datasets. 

\begin{table*}[!t]
\centering
\resizebox{\textwidth}{!}{
\begin{tabular}{ll ccccc ccccc}
\toprule
{} & {} &\multicolumn{5}{c}{\textbf{IoU Surface}} & \multicolumn{5}{c}{\textbf{IoU Bounding Box}}\\
\cmidrule(r){3-7} \cmidrule(r){8-12}
\textbf{Subject} & \textbf{Clothing} & \textbf{AVG-LBS} & \textbf{FIRST-LBS} & \textbf{SNARF} & \textbf{SNARF-NC} & \textbf{INS (ours)} & \textbf{AVG-LBS} & \textbf{FIRST-LBS} & \textbf{SNARF} & \textbf{SNARF-NC} & \textbf{INS (ours)}\\
\midrule
50002   & longshort &  \underline{84.80\%}  & 87.89\% & 47.34\%    & 96.56\%  & \textbf{97.50\%}  & 63.86\%  & 66.42\% & 85.41\%  & 84.02\%  &\textbf{89.57\%}   \\

03375 & blazerlong & 62.63\% & 53.91\% & \underline{70.16\%} & 66.1\% & \textbf{72.83\%} & 62.57\% & 54.05\% & \underline{70.24\%} & 66.11\% & \textbf{72.69\%}  \\
00215 & poloshort & 64.91\% & 57.27\% & \textbf{73.1\%} & 65.68\% & \underline{72.22\%} & 64.92\% & 57.55\% & \textbf{73.1\%} & 65.8\% & \underline{72.35\%}  \\
00096 & shirtlong & 63.48\% & 52.07\% & \textbf{75.85\%} & 67.92\% & \underline{73.48\%} & 63.55\% & 52.15\% & \textbf{75.77\%} & 67.97\% & \underline{73.56\%}  \\
00096 & shirtshort & 59.05\% & 54.67\% & \textbf{75.32\%} & 63.59\% & \underline{74.89\%} & 59.12\% & 54.7\% & \textbf{75.16\%} & 63.61\% & \underline{74.96\%}  \\
00096 & jerseyshort & 62.0\% & 56.63\% & \underline{73.28\%} & 64.28\% & \textbf{74.61\%} & 62.02\% & 56.45\% & \underline{73.19\%} & 63.98\% & \textbf{74.37\%}  \\
00134 & longlong & 65.98\% & 61.5\% & \underline{73.96\%} & 67.65\% & \textbf{78.97\%} & 65.91\% & 61.45\% & \underline{73.96\%} & 67.73\% & \textbf{78.93\%}  \\
03223 & shortshort & 74.74\% & 60.66\% & \underline{81.42\%} & 77.56\% & \textbf{82.63\%} & 74.81\% & 60.67\% & \underline{81.33\%} & 77.66\% & \textbf{82.76\%}  \\
03331 & longshort & 70.3\% & 65.2\% & \textbf{77.13\%} & 75.12\% & \underline{77.12\%} & 70.5\% & 65.52\% & \underline{77.03\%} & 74.5\% & \textbf{77.22\%}  \\
00127 & shortlong & \textbf{73.84\%} & 65.7\% & 72.48\% & 70.0\% & \underline{73.2\%} & \textbf{74.09\%} & 65.87\% & 72.31\% & 69.9\% & \underline{73.33\%}  \\
02474 & longshort & 60.87\% & 54.37\% & \underline{70.16\%} & 62.39\% & \textbf{71.64\%} & 60.94\% & 54.41\% & \underline{70.14\%} & 62.26\% & \textbf{71.76\%}  \\
03284 & longshort & 64.68\% & 58.5\% & \underline{67.04\%} & 65.76\% & \textbf{68.79\%} & 64.68\% & 58.21\% & \underline{66.87\%} & 65.44\% & \textbf{68.77\%}  \\
00032 & longshort & 64.05\% & 59.63\% & \textbf{69.23\%} & 64.47\% & \underline{69.04\%} & 64.51\% & 59.67\% & \textbf{69.21\%} & 64.63\% & \underline{69.14\%}  \\
00122 & shortlong & 56.85\% & 45.74\% & \underline{64.67\%} & 60.98\% & \textbf{64.85\%} & 57.01\% & 45.85\% & \underline{64.56\%} & 60.87\% & \textbf{65.06\%}  \\
03394 & longlong & 62.56\% & 52.11\% & \underline{69.43\%} & 66.16\% & \textbf{71.55\%} & 62.6\% & 52.25\% & \underline{69.27\%} & 66.1\% & \textbf{71.36\%}  \\
00159 & longshort & 69.27\% & 63.22\% & \underline{70.31\%} & 65.64\% & \textbf{71.1\%} & 69.52\% & 63.76\% & \underline{70.35\%} & 65.19\% & \textbf{71.58\%}  \\
\midrule
\textbf{Average} &  & 65.01\% & 57.41\% & \underline{72.24\%} & 66.89\% & \textbf{73.13\%} & 65.12\% & 57.5\% & \underline{72.17\%} & 66.78\% & \textbf{73.19\%}  \\

\bottomrule
\end{tabular}
}
\caption{\textbf{Quantitative Results on Clothed Humans.} We find our approach INS outperforms all methods when averaged across 15 runs, on both IoU Surface and IoU Bounding Box metrics.  \vspace{-1em}}
\label{tab:fcape}
\end{table*}

{\small
\bibliographystyle{ieee_fullname}
\bibliography{cvpr/references}

\begin{thebibliography}{10}\itemsep=-1pt

\bibitem{alldieck2021imghum}
Thiemo Alldieck, Hongyi Xu, and Cristian Sminchisescu.
\newblock {imGHUM}: Implicit generative models of 3d human shape and
  articulated pose.
\newblock In {\em Int. Conf. Comput. Vis.}, 2021.

\bibitem{anguelov2005scape}
Dragomir Anguelov, Praveen Srinivasan, Daphne Koller, Sebastian Thrun, Jim
  Rodgers, and James Davis.
\newblock Scape: shape completion and animation of people.
\newblock In {\em ACM SIGGRAPH 2005 Papers}, pages 408--416, 2005.

\bibitem{analyze_inv}
Lynton Ardizzone, Jakob Kruse, Sebastian Wirkert, Daniel Rahner, Eric~W.
  Pellegrini, Ralf~S. Klessen, Lena Maier-Hein, Carsten Rother, and Ullrich
  Köthe.
\newblock Analyzing inverse problems with invertible neural networks, 2018.

\bibitem{dogsout}
Benjamin Biggs, Oliver Boyne, James Charles, Andrew Fitzgibbon, and Roberto
  Cipolla.
\newblock Who left the dogs out? 3d animal reconstruction with expectation
  maximization in the loop.
\newblock In Andrea Vedaldi, Horst Bischof, Thomas Brox, and Jan-Michael Frahm,
  editors, {\em Computer Vision -- ECCV 2020}, 2020.

\bibitem{biggs2018creatures}
Benjamin Biggs, Thomas Roddick, Andrew Fitzgibbon, and Roberto Cipolla.
\newblock Creatures great and smal: Recovering the shape and motion of animals
  from video.
\newblock In {\em Asian Conference on Computer Vision}, pages 3--19. Springer,
  2018.

\bibitem{Bogo:ECCV:2016}
Federica Bogo, Angjoo Kanazawa, Christoph Lassner, Peter Gehler, Javier Romero,
  and Michael~J. Black.
\newblock Keep it {SMPL}: Automatic estimation of {3D} human pose and shape
  from a single image.
\newblock In {\em Eur. Conf. Comput. Vis.}, 2016.

\bibitem{Broyden1965BOOK}
Charles~G Broyden.
\newblock A class of methods for solving nonlinear simultaneous equations.
\newblock {\em Mathematics of computation}, pages 19(92):577--593, 1965.

\bibitem{Burov_2021_ICCV}
Andrei Burov, Matthias Nie{\ss}ner, and Justus Thies.
\newblock Dynamic surface function networks for clothed human bodies.
\newblock In {\em Proceedings of the IEEE/CVF International Conference on
  Computer Vision (ICCV)}, pages 10754--10764, October 2021.

\bibitem{Cai2022NDR}
Hongrui Cai, Wanquan Feng, Xuetao Feng, Yan Wang, and Juyong Zhang.
\newblock Neural surface reconstruction of dynamic scenes with monocular rgb-d
  camera.
\newblock In {\em Thirty-sixth Conference on Neural Information Processing
  Systems (NeurIPS)}, 2022.

\bibitem{gDNA}
Xu Chen, Tianjian Jiang, Jie Song, Jinlong Yang, Michael~J. Black, Andreas
  Geiger, and Otmar Hilliges.
\newblock gdna: Towards generative detailed neural avatars, 2022.

\bibitem{snarf}
Xu Chen, Yufeng Zheng, Michael~J Black, Otmar Hilliges, and Andreas Geiger.
\newblock Snarf: Differentiable forward skinning for animating non-rigid neural
  implicit shapes.
\newblock {\em arXiv preprint arXiv:2104.03953}, 2021.

\bibitem{SMPLicit:2021}
Enric Corona, Albert Pumarola, Guillem Aleny{\`a}, Gerard Pons-Moll, and
  Francesc Moreno-Noguer.
\newblock {SMPL}icit: Topology-aware generative model for clothed people.
\newblock In {\em Proceedings IEEE Conf. on Computer Vision and Pattern
  Recognition (CVPR)}, June 2021.

\bibitem{deng2019neural}
Boyang Deng, JP Lewis, Timothy Jeruzalski, Gerard Pons-Moll, Geoffrey Hinton,
  Mohammad Norouzi, and Andrea Tagliasacchi.
\newblock Neural articulated shape approximation.
\newblock In {\em The European Conference on Computer Vision (ECCV)}.
  {Springer}, August 2020.

\bibitem{nasa}
Boyang Deng, John~P Lewis, Timothy Jeruzalski, Gerard Pons-Moll, Geoffrey
  Hinton, Mohammad Norouzi, and Andrea Tagliasacchi.
\newblock Nasa neural articulated shape approximation.
\newblock In {\em Computer Vision--ECCV 2020: 16th European Conference,
  Glasgow, UK, August 23--28, 2020, Proceedings, Part VII 16}, pages 612--628.
  Springer, 2020.

\bibitem{nice}
Laurent Dinh, David Krueger, and Yoshua Bengio.
\newblock Nice: Non-linear independent components estimation.
\newblock {\em arXiv preprint arXiv:1410.8516}, 2014.

\bibitem{realnvp}
Laurent Dinh, Jascha Sohl-Dickstein, and Samy Bengio.
\newblock Density estimation using real nvp.
\newblock {\em arXiv preprint arXiv:1605.08803}, 2016.

\bibitem{Drobyshev22MP}
Nikita Drobyshev, Jenya Chelishev, Taras Khakhulin, Aleksei Ivakhnenko, Victor
  Lempitsky, and Egor Zakharov.
\newblock Megaportraits: One-shot megapixel neural head avatars.
\newblock In {\em Proceedings of the 30th ACM International Conference on
  Multimedia}, 2022.

\bibitem{behrmann2019invertible}
Behrmann et al.
\newblock Invertible residual networks.
\newblock In {\em ICML}, 2019.

\bibitem{sgd}
Keskar et al.
\newblock On large-batch training for deep learning: Generalization gap and
  sharp minima, 2017.

\bibitem{zheng2022IMavatar}
Zheng et al.
\newblock {I} {M} {Avatar}: Implicit morphable head avatars from videos.
\newblock In {\em CVPR}, 2022.

\bibitem{germain2015made}
Mathieu Germain, Karol Gregor, Iain Murray, and Hugo Larochelle.
\newblock Made: Masked autoencoder for distribution estimation.
\newblock In {\em International Conference on Machine Learning}, pages
  881--889. PMLR, 2015.

\bibitem{Guler_2019_CVPR}
Riza~Alp Guler and Iasonas Kokkinos.
\newblock Holopose: Holistic 3d human reconstruction in-the-wild.
\newblock In {\em Proceedings of the IEEE/CVF Conference on Computer Vision and
  Pattern Recognition (CVPR)}, June 2019.

\bibitem{huang2020arch}
Zeng Huang, Yuanlu Xu, Christoph Lassner, Hao Li, and Tony Tung.
\newblock {ARCH:} animatable reconstruction of clothed humans.
\newblock In {\em 2020 {IEEE/CVF} Conference on Computer Vision and Pattern
  Recognition, {CVPR} 2020, Seattle, WA, USA, June 13-19, 2020}, pages
  3090--3099. {IEEE}, 2020.

\bibitem{irevnet}
Jörn-Henrik Jacobsen, Arnold~W.M. Smeulders, and Edouard Oyallon.
\newblock i-revnet: Deep invertible networks.
\newblock In {\em International Conference on Learning Representations}, 2018.

\bibitem{kingma2014adam}
Diederik~P Kingma and Jimmy Ba.
\newblock Adam: {A} method for stochastic optimization.
\newblock In {\em Int. Conf. Learn. Represent.}, 2015.

\bibitem{kingma2018glow}
Diederik~P Kingma and Prafulla Dhariwal.
\newblock Glow: Generative flow with invertible 1x1 convolutions.
\newblock {\em arXiv preprint arXiv:1807.03039}, 2018.

\bibitem{Kobyzev_2021}
Ivan Kobyzev, Simon~J.D. Prince, and Marcus~A. Brubaker.
\newblock Normalizing flows: An introduction and review of current methods.
\newblock {\em {IEEE} Transactions on Pattern Analysis and Machine
  Intelligence}, 2021.

\bibitem{Lei2022CaDeX}
Jiahui Lei and Kostas Daniilidis.
\newblock Cadex: Learning canonical deformation coordinate space for dynamic
  surface representation via neural homeomorphism.
\newblock In {\em Proceedings of the IEEE/CVF Conference on Computer Vision and
  Pattern Recognition}, 2022.

\bibitem{tava}
Ruilong Li, Julian Tanke, Minh Vo, Michael Zollhofer, Jurgen Gall, Angjoo
  Kanazawa, and Christoph Lassner.
\newblock Tava: Template-free animatable volumetric actors, 2022.

\bibitem{FLAME:SiggraphAsia2017}
Tianye Li, Timo Bolkart, Michael.~J. Black, Hao Li, and Javier Romero.
\newblock Learning a model of facial shape and expression from {4D} scans.
\newblock {\em ACM Transactions on Graphics, (Proc. SIGGRAPH Asia)},
  36(6):194:1--194:17, 2017.

\bibitem{SMPL:2015}
Matthew Loper, Naureen Mahmood, Javier Romero, Gerard Pons-Moll, and Michael~J.
  Black.
\newblock {SMPL}: A skinned multi-person linear model.
\newblock {\em ACM Trans. Graphics (Proc. SIGGRAPH Asia)}, 34(6):248:1--248:16,
  Oct. 2015.

\bibitem{lorensen1987marching}
William~E. Lorensen and Harvey~E. Cline.
\newblock Marching cubes: {A} high resolution 3{D} surface construction
  algorithm.
\newblock In {\em Proceedings of the 14th Annual Conference on Computer
  Graphics and Interactive Techniques, {SIGGRAPH} 1987, Anaheim, California,
  USA, July 27-31, 1987}, pages 163--169. {ACM}, 1987.

\bibitem{SkiRT:3DV:2022}
Qianli Ma, Jinlong Yang, Michael~J. Black, and Siyu Tang.
\newblock Neural point-based shape modeling of humans in challenging clothing.
\newblock In {\em 2022 International Conference on 3D Vision (3DV)}, September
  2022.

\bibitem{CAPE:CVPR:20}
Qianli Ma, Jinlong Yang, Anurag Ranjan, Sergi Pujades, Gerard Pons{-}Moll, Siyu
  Tang, and Michael~J. Black.
\newblock Learning to dress {3}d people in generative clothing.
\newblock In {\em 2020 {IEEE/CVF} Conference on Computer Vision and Pattern
  Recognition, {CVPR} 2020, Seattle, WA, USA, June 13-19, 2020}, pages
  6468--6477. {IEEE}, 2020.

\bibitem{AMASS:ICCV:2019}
Naureen Mahmood, Nima Ghorbani, Nikolaus~F. Troje, Gerard Pons-Moll, and
  Michael~J. Black.
\newblock {AMASS}: Archive of motion capture as surface shapes.
\newblock In {\em Int. Conf. Comput. Vis.}, 2019.

\bibitem{mescheder2019occupancy}
Lars~M. Mescheder, Michael Oechsle, Michael Niemeyer, Sebastian Nowozin, and
  Andreas Geiger.
\newblock Occupancy networks: Learning 3{D} reconstruction in function space.
\newblock In {\em {IEEE} Conference on Computer Vision and Pattern Recognition,
  {CVPR} 2019, Long Beach, CA, USA, June 16-20, 2019}, pages 4460--4470.
  Computer Vision Foundation / {IEEE}, 2019.

\bibitem{LEAP:CVPR:21}
Marko Mihajlovic, Yan Zhang, Michael~J. Black, and Siyu Tang.
\newblock {LEAP}: Learning articulated occupancy of people.
\newblock In {\em Proceedings IEEE Conf. on Computer Vision and Pattern
  Recognition (CVPR)}, June 2021.

\bibitem{mildenhall2020nerf}
Ben Mildenhall, Pratul~P Srinivasan, Matthew Tancik, Jonathan~T Barron, Ravi
  Ramamoorthi, and Ren Ng.
\newblock Nerf: Representing scenes as neural radiance fields for view
  synthesis.
\newblock In {\em European conference on computer vision}, pages 405--421.
  Springer, 2020.

\bibitem{moon2020deephandmesh}
Gyeongsik Moon, Takaaki Shiratori, and Kyoung~Mu Lee.
\newblock Deephandmesh: {A} weakly-supervised deep encoder-decoder framework
  for high-fidelity hand mesh modeling.
\newblock In {\em Computer Vision - {ECCV} 2020 - 16th European Conference,
  Glasgow, UK, August 23-28, 2020, Proceedings, Part {II}}, volume 12347 of
  {\em Lecture Notes in Computer Science}, pages 440--455. Springer, 2020.

\bibitem{osman2020star}
Ahmed A.~A. Osman, Timo Bolkart, and Michael~J. Black.
\newblock {STAR:} sparse trained articulated human body regressor.
\newblock In {\em Computer Vision - {ECCV} 2020 - 16th European Conference,
  Glasgow, UK, August 23-28, 2020, Proceedings, Part {VI}}, volume 12351 of
  {\em Lecture Notes in Computer Science}, pages 598--613. Springer, 2020.

\bibitem{papamakarios2017masked}
George Papamakarios, Theo Pavlakou, and Iain Murray.
\newblock Masked autoregressive flow for density estimation.
\newblock {\em arXiv preprint arXiv:1705.07057}, 2017.

\bibitem{nerfies}
Keunhong Park, Utkarsh Sinha, Jonathan~T. Barron, Sofien Bouaziz, Dan~B
  Goldman, Steven~M. Seitz, and Ricardo Martin-Brualla.
\newblock Nerfies: Deformable neural radiance fields, 2020.

\bibitem{neuralparts}
Despoina Paschalidou, Angelos Katharopoulos, Andreas Geiger, and Sanja Fidler.
\newblock Neural parts: Learning expressive 3d shape abstractions with
  invertible neural networks.
\newblock In {\em Proceedings of the IEEE/CVF Conference on Computer Vision and
  Pattern Recognition}, pages 3204--3215, 2021.

\bibitem{Paszke_PyTorch_An_Imperative_2019}
Adam Paszke, Sam Gross, Francisco Massa, Adam Lerer, James Bradbury, Gregory
  Chanan, Trevor Killeen, Zeming Lin, Natalia Gimelshein, Luca Antiga, Alban
  Desmaison, Andreas Kopf, Edward Yang, Zachary DeVito, Martin Raison, Alykhan
  Tejani, Sasank Chilamkurthy, Benoit Steiner, Lu Fang, Junjie Bai, and Soumith
  Chintala.
\newblock {PyTorch: An Imperative Style, High-Performance Deep Learning
  Library}.
\newblock In {\em Advances in Neural Information Processing Systems 32}, pages
  8024--8035. Curran Associates, Inc., 2019.

\bibitem{SMPL-X:2019}
Georgios Pavlakos, Vasileios Choutas, Nima Ghorbani, Timo Bolkart, Ahmed A.~A.
  Osman, Dimitrios Tzionas, and Michael~J. Black.
\newblock Expressive body capture: 3d hands, face, and body from a single
  image.
\newblock In {\em Proceedings IEEE Conf. on Computer Vision and Pattern
  Recognition (CVPR)}, 2019.

\bibitem{peng2021animatable}
Sida Peng, Junting Dong, Qianqian Wang, Shangzhan Zhang, Qing Shuai, Hujun Bao,
  and Xiaowei Zhou.
\newblock Animatable neural radiance fields for human body modeling.
\newblock {\em arXiv preprint arXiv:2105.02872}, 2021.

\bibitem{neuralbody}
Sida Peng, Yuanqing Zhang, Yinghao Xu, Qianqian Wang, Qing Shuai, Hujun Bao,
  and Xiaowei Zhou.
\newblock Neural body: Implicit neural representations with structured latent
  codes for novel view synthesis of dynamic humans.
\newblock In {\em Proceedings of the IEEE/CVF Conference on Computer Vision and
  Pattern Recognition}, pages 9054--9063, 2021.

\bibitem{pinn}
Chengping Rao, Hao Sun, and Yang Liu.
\newblock Physics informed deep learning for computational elastodynamics
  without labeled data, 2020.

\bibitem{romero2017embodied}
Javier Romero, Dimitrios Tzionas, and Michael~J Black.
\newblock Embodied hands: Modeling and capturing hands and bodies together.
\newblock {\em ACM Transactions on Graphics (ToG)}, 36(6):1--17, 2017.

\bibitem{saito2019pifu}
Shunsuke Saito, Zeng Huang, Ryota Natsume, Shigeo Morishima, Hao Li, and Angjoo
  Kanazawa.
\newblock {PIF}u: Pixel-aligned implicit function for high-resolution clothed
  human digitization.
\newblock In {\em 2019 {IEEE/CVF} International Conference on Computer Vision,
  {ICCV} 2019, Seoul, Korea (South), October 27 - November 2, 2019}, pages
  2304--2314. {IEEE}, 2019.

\bibitem{saito2020pifuhd}
Shunsuke Saito, Tomas Simon, Jason~M. Saragih, and Hanbyul Joo.
\newblock {PIF}u{HD}: Multi-level pixel-aligned implicit function for
  high-resolution 3{D} human digitization.
\newblock In {\em 2020 {IEEE/CVF} Conference on Computer Vision and Pattern
  Recognition, {CVPR} 2020, Seattle, WA, USA, June 13-19, 2020}, pages 81--90.
  {IEEE}, 2020.

\bibitem{scanimate}
Shunsuke Saito, Jinlong Yang, Qianli Ma, and Michael~J Black.
\newblock Scanimate: Weakly supervised learning of skinned clothed avatar
  networks.
\newblock In {\em Proceedings of the IEEE/CVF Conference on Computer Vision and
  Pattern Recognition}, pages 2886--2897, 2021.

\bibitem{shao2022doublefield}
Ruizhi Shao, Hongwen Zhang, He Zhang, Mingjia Chen, Yanpei Cao, Tao Yu, and
  Yebin Liu.
\newblock Doublefield: Bridging the neural surface and radiance fields for
  high-fidelity human reconstruction and rendering.
\newblock In {\em CVPR}, 2022.

\bibitem{sitzmann2019siren}
Vincent Sitzmann, Julien N.~P. Martel, Alexander~W. Bergman, David~B. Lindell,
  and Gordon Wetzstein.
\newblock Implicit neural representations with periodic activation functions.
\newblock In {\em Advances in Neural Information Processing Systems 33: Annual
  Conference on Neural Information Processing Systems 2020, NeurIPS 2020,
  December 6-12, 2020, virtual}, 2020.

\bibitem{Su2022DANBODA}
Shih-Yang Su, Timur~M. Bagautdinov, and Helge Rhodin.
\newblock Danbo: Disentangled articulated neural body representations via graph
  neural networks.
\newblock {\em ArXiv}, abs/2205.01666, 2022.

\bibitem{metaavatar}
Shaofei Wang, Marko Mihajlovic, Qianli Ma, Andreas Geiger, and Siyu Tang.
\newblock Metaavatar: Learning animatable clothed human models from few depth
  images.
\newblock {\em arXiv preprint arXiv:2106.11944}, 2021.

\bibitem{ARAH:ECCV:2022}
Shaofei Wang, Katja Schwarz, Andreas Geiger, and Siyu Tang.
\newblock Arah: Animatable volume rendering of articulated human sdfs.
\newblock In {\em European Conference on Computer Vision}, 2022.

\bibitem{human-nerf}
Chung-Yi Weng, Brian Curless, Pratul~P. Srinivasan, Jonathan~T. Barron, and Ira
  Kemelmacher-Shlizerman.
\newblock Humannerf: Free-viewpoint rendering of moving people from monocular
  video, 2022.

\bibitem{xiu2022icon}
Yuliang Xiu, Jinlong Yang, Dimitrios Tzionas, and Michael~J. Black.
\newblock {ICON}: {I}mplicit {C}lothed humans {O}btained from {N}ormals.
\newblock In {\em Proceedings of the IEEE/CVF Conference on Computer Vision and
  Pattern Recognition (CVPR)}, pages 13296--13306, June 2022.

\bibitem{hnerf}
Hongyi Xu, Thiemo Alldieck, and Cristian Sminchisescu.
\newblock H-nerf: Neural radiance fields for rendering and temporal
  reconstruction of humans in motion, 2021.

\bibitem{xu2020ghum}
Hongyi Xu, Eduard~Gabriel Bazavan, Andrei Zanfir, William~T. Freeman, Rahul
  Sukthankar, and Cristian Sminchisescu.
\newblock {GHUM} {\&} {GHUML:} generative 3{D} human shape and articulated pose
  models.
\newblock In {\em 2020 {IEEE/CVF} Conference on Computer Vision and Pattern
  Recognition, {CVPR} 2020, Seattle, WA, USA, June 13-19, 2020}, pages
  6183--6192. {IEEE}, 2020.

\bibitem{yang2022banmo}
Gengshan Yang, Minh Vo, Natalia Neverova, Deva Ramanan, Andrea Vedaldi, and
  Hanbyul Joo.
\newblock Banmo: Building animatable 3d neural models from many casual videos.
\newblock In {\em Proceedings of the IEEE/CVF Conference on Computer Vision and
  Pattern Recognition}, pages 2863--2873, 2022.

\bibitem{zhao2022avatar}
Hao Zhao, Jinsong Zhang, Yu-Kun Lai, Zerong Zheng, Yingdi Xie, Yebin Liu, and
  Kun Li.
\newblock High-fidelity human avatars from a single rgb camera.
\newblock In {\em CVPR}, 2022.

\bibitem{slrf}
Zerong Zheng, Han Huang, Tao Yu, Hongwen Zhang, Yandong Guo, and Yebin Liu.
\newblock Structured local radiance fields for human avatar modeling, 2022.

\bibitem{Zuffi_2019_ICCV}
Silvia Zuffi, Angjoo Kanazawa, Tanya Berger-Wolf, and Michael~J. Black.
\newblock Three-d safari: Learning to estimate zebra pose, shape, and texture
  from images "in the wild".
\newblock In {\em Proceedings of the IEEE/CVF International Conference on
  Computer Vision (ICCV)}, October 2019.

\end{thebibliography}
}

\end{document}